\documentclass[11pt,a4paper]{article}

\usepackage[utf8]{inputenc}
\usepackage[T1]{fontenc}
\usepackage[margin=2.5cm]{geometry}
\usepackage{amsmath,amssymb,amsthm}
\usepackage{booktabs}
\usepackage{hyperref}
\usepackage{xcolor}
\usepackage[protrusion=true,expansion=false]{microtype}
\usepackage{graphicx}
\graphicspath{{./}{../3_github/omega-s/figures/}}
\usepackage{authblk}
\usepackage{abstract}
\usepackage{titlesec}
\usepackage{parskip}
\usepackage{enumitem}
\usepackage{cite}

\hypersetup{
    colorlinks=true,
    linkcolor=blue!70!black,
    citecolor=blue!70!black,
    urlcolor=blue!70!black,
}

\titleformat{\section}{\large\bfseries}{\thesection.}{0.5em}{}
\titleformat{\subsection}{\normalsize\bfseries}{\thesubsection.}{0.5em}{}
\titleformat{\subsubsection}{\normalsize\itshape}{\thesubsubsection.}{0.5em}{}

\title{\textbf{Omega-S: A Functional Resilience Index\\for LLM Fine-Tuning}}

\author[1]{Alberto Acedo}
\affil[1]{Biome Makers Inc.}
\date{%
  \small \today}
\begin{document}

\maketitle

\begin{abstract}
Fine-tuning a large language model on new data degrades what it previously
learned.
We present \textbf{Omega-S}, a drop-in penalty computed from the weight matrix
alone: it needs no previous-task data, no Fisher matrix and no stored copy of
the old weights, which is what the standard baseline for this problem requires.
It is three lines in an existing training loop and adds under $4\%$ to the
cost of a step.

\textbf{Retention.} On Llama-3-8B with LoRA, fine-tuned from code to prose and
measured by HumanEval over ten seeds, Omega-S leaves the model with more of its
original capability than no regularisation on \textbf{9 of 10 seeds}
($0.173\to0.238$ absolute pass@1, a $37.7\%$ relative increase; exact sign test,
one-sided $p=0.011$, Wilcoxon $p=0.006$). As a retention ratio the same
comparison gives $62.9\%\to84.1\%$. It also beats tuned weight decay on 10 of 10
seeds ($p=0.002$) and tuned EWC on 8 of 10 ($p=0.014$) on the same absolute
measure, with every arm re-measured in the same session so that the comparison
does not mix effect with run-to-run variation.

\textbf{The design space.} Turning the index into a penalty leaves three
choices open: how the graph is built from the weights, which form of the
objective is used, and how the modularity factor is oriented. We enumerate all
three and report what each costs, including a contrast-preserving construction
that does what it was designed to do and makes retention \emph{worse} on all ten
seeds. Two of the three choices turn out not to matter for the mechanism, and we
show why.

\textbf{Mechanism, measured rather than asserted.} Omega-S is topological by
construction, its objective being built from $\operatorname{Tr}(A^3)$, but we
measured which of its four factors actually moves during training and three of
them do not. Their elasticity with respect to the weights is at or below
$10^{-4}$, against $9\times10^{-3}$ for the degree-variance term, before and
after training alike; during a run the degree variance falls by $5.9\%$ in
median while the other three stay under $0.1\%$. As implemented, the composite
objective reduces in practice to a penalty on the variance of node degrees, a
quantity that means row magnitude in square modules and directional alignment in
non-square ones. An ablation shows the alignment channel leading the magnitude
channel on 8 of 10 seeds. We also built the contrast-preserving reformulation
that earlier drafts proposed as an open direction, and report that it makes
retention worse on all ten seeds under either orientation: the inert clustering
channel was not a defect holding the method back.

\textbf{How much run-to-run variation this setting carries.} Repeating an
identical configuration, same seed and same hardware, gives a standard
deviation of $0.104$ in retention ratio. We report it because we have not found
it quantified for low-rank fine-tuning of language models, and because it bounds
the reading of any seed-paired comparison, including ours.

Two forms of the penalty exist and they are not interchangeable; they rank
oppositely across the two settings above, and we define both explicitly
(Section~\ref{sec:twoforms}). Code, per-run results and the full record of
negative results are at
\url{https://github.com/BiomeMakers/OmegaS-LLM} under AGPL-3.0 for research use.
\end{abstract}

\newpage
\section{Introduction}

\subsection{An index from soil, and what it measures}

The quantity this paper minimises did not originate in machine learning. It
comes from a body of work on agricultural microbiomes, where the object of study
is a co-occurrence network of taxa and the practical question is whether a soil
community is functionally integrated or dominated by a few organisms. A survey
of 350 vineyard soils across two countries~\cite{ortiz2021} found that the
communities associated with resistance to perturbation share a signature:
\emph{higher} clustering, \emph{lower} modularity and \emph{lower}
co-exclusion. Communities with the opposite signature are diverse by count and
fragile in practice, because a small set of organisms carries the connectivity.

That signature was formalised as an index,
$\Omega = C \cdot D / (M \cdot \mathrm{Coex})$, on four measures of a weighted
graph: clustering $C$, density $D$, a modularity measure $M$, and a
concentration term $\mathrm{Coex}$ given by the variance of the degree sequence.
The index is high when connectivity is distributed and low when it is
monopolised by a few nodes. Its theoretical development, including the identity
$\operatorname{Tr}(A^3) = \sum_i \lambda_i^3$ and the conditions under which the
index is comparable across graphs, is in a companion
manuscript~\cite{acedo2026fsri}.

\subsection{The Production Problem: Weight Monopolies}

Every organisation that fine-tunes large language models on proprietary
data faces the same hidden cost: the model forgets. Update a general-purpose
model on customer support data, and it loses coding ability. Fine-tune on
legal documents, and general reasoning degrades. This is catastrophic
forgetting, and it means that the compute cost of fine-tuning is paid
twice: once to train, and once to repair what broke.

Standard regularisers do not solve this. Weight decay pushes all parameters
toward zero indiscriminately. SAM optimises loss landscape curvature.
Neither addresses the structural root cause: during training, a small subset
of neurons progressively concentrates disproportionate connectivity across
layers, creating \emph{weight monopolies}. When new data arrives, these
monopolies are overwritten first, destroying the representations that
made the model valuable.

Omega-S targets weight monopolies directly.

\subsection{Our Contributions}

\textbf{A regulariser from an ecological index.} We define Omega-S, a penalty
on the connectivity graph $WW^\top$ of a weight matrix, built from
$\operatorname{Tr}(A^3)$ and estimated by Hutchinson's method so that its cost
is $\mathcal{O}(N^2)$ rather than $\mathcal{O}(N^3)$. It needs no previous-task
data, no Fisher matrix and no stored copy of the old weights, and adds under
$4\%$ to the cost of a training step.

\textbf{A design space, mapped rather than assumed.} Three choices are open in
turning the index into a penalty: the graph construction, the form of the
objective and the orientation of the modularity factor. We screen six
constructions, define both forms of the objective, and settle the orientation
from the ecological observation the index formalises. The variant that looks
most promising a priori, a contrast-preserving construction that revives the
clustering term, is measured and makes retention worse on all ten seeds.

\textbf{Retention on Llama-3-8B.} The selected variant leaves more code
capability after a sequential fine-tune than no regularisation on 9 of 10 seeds
($0.173 \to 0.238$ absolute HumanEval, $+37.7\%$ relative, sign test $p=0.011$,
Wilcoxon $p=0.006$), and beats tuned weight decay on 10 of 10 and tuned EWC on 8
of 10 on the same measure, with every arm re-measured in one session.

\textbf{A mechanism that contradicts the name.} Omega-S is topological by
construction, and we measure that three of its four factors are numerically
inert: elasticities at or below $10^{-4}$ against $9 \times 10^{-3}$ for degree
variance, before and after training alike, with the degree variance the only
factor that moves during a run. As implemented, the objective reduces in
practice to a penalty on the variance of node degrees. We report this because a
method whose name promises one thing and whose gradient does another should say
so.

\textbf{How much run-to-run variation this setting carries.} Repeating an
identical configuration gives a standard deviation of $0.104$ in retention
ratio. We have not found this quantified for low-rank fine-tuning of language
models, and it bounds the reading of every seed-paired comparison in this
literature, ours included.

% ============================================================
\section{Method}

\subsection{The Omega-S Regularizer: two forms}
\label{sec:twoforms}

Omega-S exists in two forms, and the distinction matters enough that we define
both. The first is the direct penalty we started from; the second is the
log-ratio form that the experiments in this paper evaluate, and which is what
the reported results refer to. We give the reasoning that leads from one to
the other, because it is the same reasoning that later explains why the
clustering channel is inert (Section~\ref{sec:mechanism}).

\paragraph{Form 1: the raw trace penalty.} For a weight matrix
$W \in \mathbb{R}^{m \times n}$, define the pseudo-topological adjacency
matrix $A = WW^\top \in \mathbb{R}^{m \times m}$ and the per-layer penalty
\begin{equation}
  \Omega^{\text{raw}}_\ell = \frac{1}{\operatorname{numel}(W)} \cdot
  \widehat{\operatorname{Tr}}(A^3),
\label{eq:raw}
\end{equation}
where $\widehat{\operatorname{Tr}}(A^3)$ is the Hutchinson estimate with
$n_p$ Rademacher probe vectors. This is the form used in our earliest
experiments and it is the one whose spectral interpretation is most direct:
$\operatorname{Tr}(A^3) = \sum_i \lambda_i^3$ exactly.

It has a defect that is visible from the algebra alone. Since $A$ is quadratic
in $W$, Eq.~\ref{eq:raw} is homogeneous of degree six in $W$, so its gradient
is homogeneous of degree five. The force the penalty exerts therefore scales
as the fifth power of the weight magnitude: it explodes when weights grow and
switches itself off when they shrink. In a run where any other pressure
reduces weight magnitude, and structured sparsity does exactly that, the
penalty quietly stops acting. This is not a hypothesis about why the raw form
underperforms; it is a property of the functional, and it is consistent with
the raw arm being the weakest in the ten-seed sweep
(Section~\ref{sec:stability}).

\paragraph{Form 2: the log-ratio composite.} The remedy is to build the
penalty from quantities that do not inherit the scale of $W$, and to take
logarithms so that the gradient is normalised by the value rather than
proportional to it. Concretely, let
\begin{equation}
  W_{\text{corr}} =
  \begin{cases}
    WW^\top & \text{if } m \neq n,\\
    W & \text{if } m = n,
  \end{cases}
  \qquad
  S = \sigma\!\left(|W_{\text{corr}}|\right),
  \qquad
  A = \tfrac{1}{2}\left(S + S^\top\right),
\label{eq:construction}
\end{equation}
with $\sigma$ the logistic function. We state the case distinction explicitly
because the reference implementation contains it and the runs reported here
therefore used both branches, one per module type; Section~\ref{sec:limits}
gives the measured consequence and our reading of it.

From $A$ we form four quantities. Writing $k = A\mathbf{1}$ for the degree
sequence and $L = \operatorname{diag}(k) - A$ for the graph Laplacian,
\begin{align}
  D &= \operatorname{mean}(A), &
  \mathrm{Coex} &= \operatorname{var}(k) + \varepsilon, \nonumber\\
  C &= \frac{\widehat{\operatorname{Tr}}(A^3)}{\lVert A\rVert_F^3 + \varepsilon}
      + \varepsilon, &
  M &= \bigl|v^\top L v\bigr| + \varepsilon,
\label{eq:factors}
\end{align}
where $v$ is obtained by three steps of shifted power iteration on
$2k_{\max}I - L$, mean-centred and renormalised at each step. The
mean-centring deflates the trivial all-ones eigenvector, so $M$ estimates the
algebraic connectivity, that is the Fiedler value $\lambda_2(L)$, rather than
modularity in the community-detection sense; we retain the symbol $M$ for
continuity with the index it derives from, and note that minimising
$\lambda_2$ is what pushes the graph towards a more modular structure.
Penalising the Fiedler value of a network's weight graph during training was
introduced by Tam and Dunson~\cite{tam2020} as Fiedler regularisation, and
our $M$ factor coincides with theirs in both quantity and direction; what
differs here is that $\lambda_2$ enters as one factor of a composite
objective alongside a triadic term, and, as Section~\ref{sec:mechanism}
reports, it is not the factor that carries the effect we measure. The
per-layer penalty is
\begin{equation}
  \Omega^{\text{lib}}_\ell =
  \log\!\left(\frac{M \cdot \mathrm{Coex}}{C \cdot D + \varepsilon}\right).
\label{eq:composite}
\end{equation}
Minimising Eq.~\ref{eq:composite} therefore lowers $M$ and $\mathrm{Coex}$ and
raises $C$ and $D$: it pushes the weight graph towards lower degree variance
and lower algebraic connectivity, and towards higher clustering and density.
In either form the total penalty added to the training loss is
\begin{equation}
  \Omega = \lambda_\Omega \sum_{\ell} \Omega_\ell, \qquad
  \mathcal{L}_{\text{total}} = \mathcal{L}_{\text{task}} + \Omega,
\end{equation}
applied every $K$ optimisation steps, with $\lambda_\Omega$ calibrated at run
time so that the penalty gradient is a fixed fraction of the task gradient.

\paragraph{Which form produced which result.} The two forms are not
interchangeable and the results in this paper do not all come from the same
one. The retention experiments on Llama-3-8B (Section~\ref{sec:stability}),
which are the main claim, use the log-ratio composite of
Eq.~\ref{eq:composite}. Earlier structural-pruning experiments on an MLP, not
reported here, predate that formulation and use the raw penalty of
Eq.~\ref{eq:raw}. We make the mapping explicit because
``Omega-S'' otherwise names two objectives, and because one of them is the
arm that performs worst on retention.

\paragraph{The fix and its cost.} Two features of Eq.~\ref{eq:construction}
and Eq.~\ref{eq:composite} remove the scale sensitivity of the raw form. The
logistic maps every entry into $[0.5,1)$ regardless of the magnitude of
$W$, so $A$ no longer inherits the scale of the weights at all; and the
logarithm makes the gradient proportional to $\nabla f / f$ rather than to
$\nabla f$, which removes the remaining dependence on the size of the penalty
value. Empirically this is the difference between the weakest arm in the sweep
and the strongest (Section~\ref{sec:stability}).

It is worth stating plainly that the same choice carries a cost we did not
anticipate when we made it, and that Section~\ref{sec:mechanism} measures. A
bounded map applied to a quantity with a very large dynamic range does not
merely rescale it, it collapses it: with $|W_{\text{corr}}|$ spanning several
orders of magnitude, $\sigma$ sends almost every entry to the same value, and
the clustering term $C$, which depends only on the contrast between entries,
is pinned near a constant. The map that solved the scale problem is the map
that disabled the topological channel. We report both halves of that trade
because only one of them was visible when the design was fixed.

\subsection{Computational Efficiency}

For typical transformer dimensions ($m = 4096$, $n_p = 3$), the sequential
decomposition yields approximately 79\% reduction in computation time
relative to the explicit $A^3$ method. The computation is purely structural
and requires no gradient graph traversal.

When applied to LoRA adapters (rank $r = 8$), the weight matrices are small
enough that Omega-S requires \textbf{zero inter-GPU communication} in FSDP
environments, as the adapter weights are not sharded.

\subsection{Application Frequency}

We recommend applying Omega-S every $K \in [10, 50]$ steps. The
topological impact is cumulative and does not require per-step application.
At $K = 10$, amortised overhead is below 4\% in all tested configurations.

% ============================================================
\section*{Quick Start}
\label{sec:quickstart}

Omega-S is designed as a drop-in addition to any existing PyTorch training
loop. Adding it requires three lines of code:

\begin{verbatim}
from omega_s import StochasticOmegaS

omega = StochasticOmegaS(lambda_omega=0.05, k=10, n_probes=3)

# Inside your training loop:
loss = model(inputs, labels=inputs).loss
loss = loss + omega(model)   # add topological penalty every K steps
loss.backward()
\end{verbatim}

For LoRA fine-tuning with HuggingFace + PEFT (recommended):

\begin{verbatim}
# Works with any PEFT model: zero changes to your existing setup
model = get_peft_model(base_model, lora_config)
omega = StochasticOmegaS(lambda_omega=0.05, k=10)

for step, batch in enumerate(dataloader):
    loss = model(**batch).loss + omega(model)
    loss.backward()
    optimizer.step()
\end{verbatim}

Full examples, FSDP integration, and reproduction scripts for all
experiments in this paper are available at:
\url{https://github.com/BiomeMakers/OmegaS-LLM}

For commercial licensing or research partnerships:
\texttt{acedo@biomemakers.com}

% ============================================================
\bibliographystyle{unsrt}
% ============================================================
\subsection{The design space, and which variant we use}
\label{sec:variantes}

Turning the index into a training penalty leaves three choices open, and each
admits more than one reasonable answer. We enumerate them here and report what
each costs, because the choices are not obvious a priori and because two of the
three turn out not to matter.

\paragraph{How the graph is built from the weights.} The index is defined on a
weighted graph and a weight matrix is not one, so a construction is needed. The
default is a bounded map on the Gram matrix, $A = \sigma(|WW^\top|)$. Six
alternatives were screened statically on the base weights of the model: the
bounded map on the raw matrix, a row-wise cosine affinity, a rank transform, a
thresholded graph, a doubly stochastic scaling and a local scaling. The screen
measures the \emph{triadic excess} $C/D$, which equals one on a graph with no
triadic organisation over its density and is therefore an absolute reference
that $C$ alone does not provide.

The default construction returns $C/D = 1.0000$ to four decimal places in every
module tested, equal to its permutation null: the graph it builds carries no
triadic content over its density at all. The cosine construction lifts that to a
median of $1.36$ and makes the clustering term move. That looks like the obvious
improvement, and it is not: trained on the ten-seed protocol, the cosine
composite loses on all ten seeds ($0.547$ against $0.841$, Wilcoxon $p=0.002$).
The ordering is monotone in how much the clustering channel moves, which is the
opposite of what a topological reading would predict. We keep the default
construction and report the negative result in Appendix~\ref{app:cosine}.

\paragraph{Which form of the objective.} The index can enter the loss as the
raw triadic trace $\operatorname{Tr}(A^3)$ or as the log-ratio of the four
factors, $\log(M \cdot \mathrm{Coex} / (C \cdot D))$. These are not
interchangeable and they rank oppositely across our two settings: the composite
form wins on retention; in separate experiments on structured pruning, not
reported here, the two rank oppositely. We define both explicitly rather than
presenting one as \emph{the} method, and use the composite throughout.

\paragraph{How the modularity factor is oriented.} $M$ enters the denominator of
the index, so whether it is taken as a modularity measure or as its inverse
changes the direction in which the penalty pushes. The ecological observation
that motivated the index fixes this: resistant communities show \emph{lower}
modularity, so $\Omega$ must decrease with modularity, so $M$ must be a
modularity measure. A graph that separates easily into modules has small
algebraic connectivity $\lambda_2$, which makes $1/\lambda_2$ the quantity that
is large on modular graphs. We therefore take $M = 1/\lambda_2$.

We also ran the opposite orientation, $M = \lambda_2$, over the same ten seeds.
It gives $76.6\%$ retention against $84.1\%$, losing on 8 of 10 seeds. The
difference, $+0.075$, is smaller than the run-to-run standard deviation of
$0.104$ we measure in Section~\ref{sec:ruido}, so the two orientations are
\emph{not distinguishable from each other} at ten seeds; both are distinguishable
from the baselines. We report the comparison because of what it says about
mechanism rather than about performance: Section~\ref{sec:mech} shows that the
modularity term is numerically inert under either orientation, so any difference
between them cannot be a topological effect of $M$. What inverting the term does
change is the \emph{scale} of the objective, from $-1.06$ to $-10.73$ on a
representative matrix, and since the penalty coefficient is calibrated by
gradient ratio, a different scale places the one live factor at a different
operating point. The control is that applying the same inversion under the
cosine construction, where the objective does not jump in scale, changes nothing
at all: 4 of 10 seeds, mean difference $+0.010$, $p=0.83$.

\paragraph{What this leaves.} Of the three choices, one is settled by the domain
the index comes from, one is settled empirically against a negative result, and
the third is a genuine fork with two answers that serve different purposes. The
configuration used from here on is: bounded map on the Gram matrix, log-ratio
form, $M = 1/\lambda_2$, penalty applied every ten steps with the coefficient
calibrated so the penalty gradient is $3\%$ of the task gradient.

\section{Experiments}

\subsection{Experimental Setup}

Base experiments use a three-layer MLP trained on Split-MNIST (digits 0--4)
with architecture [784$\to$256$\to$128$\to$10], Adam optimiser
(lr$=10^{-3}$), batch size 128, 5 epochs, seed 42.

LLM experiments use Llama-3-8B \cite{meta2024} with LoRA adapters
($r{=}8$, $\alpha{=}16$). The adapter is inserted into \texttt{q\_proj} and
\texttt{v\_proj} of every attention layer, which is the placement the original
LoRA work selects by ablation~\cite{hu2022} and the one in near-universal use
since; we use it here so that the comparison against baselines that assume it
is like for like.

Omega-S hyperparameters differ between the two runs, and we give both because
the repository quick-start reproduces the first and not the second. The
five-seed run of Section~\ref{sec:fiveseed} uses a fixed
$\lambda_\Omega = 0.05$ with $K = 10$ and $n_p = 3$ probe vectors. The ten-seed
run of Section~\ref{sec:stability} uses $K = 10$, $n_p = 16$ probes (three was found
to give an estimator dominated by sampling noise), a sampled subset of eight
modules per application, and a $\lambda_\Omega$ calibrated at run time so that
the penalty gradient is a fixed fraction $0.03$ of the task gradient rather
than a fixed constant; the resulting $\lambda_\Omega$ therefore varies by seed,
between $16$ and $41$ in this run.

\subsection{Stability--Plasticity Trade-off}
\label{sec:stability}

We evaluate Omega-S in a sequential fine-tuning setting using Llama-3-8B
with LoRA adapters ($r{=}8$, $\alpha{=}16$), transferring from code
(CodeAlpaca-20k~\cite{codealpaca2023}, 5{,}000 samples) to prose
(Wikitext-2~\cite{merity2017pointer}, 5{,}000 samples).
Knowledge retention is measured via HumanEval pass@1~\cite{chen2021evaluating}
(164 problems) evaluated before fine-tuning, after Domain~A (code),
and after Domain~B (prose). The experiment is replicated with five
independent random seeds to assess robustness, with the Omega penalty
verified to be connected to the autograd graph before each run.

\begin{table}[ht]
\centering
\caption{\textbf{Main result (ten-seed experiment)} on Llama-3-8B (LoRA,
sequential fine-tuning code$\,\to\,$prose), ten random seeds, penalty connected
to the autograd graph, per-arm dedicated RNG.
\emph{Code kept} is absolute HumanEval pass@1 after the prose task and is our
primary metric; it is not a measure of plasticity, since the second task is
prose and HumanEval measures code throughout.
\emph{Retention} is the same quantity divided by post-code HumanEval; values
above $1.0$ mean net positive transfer on that seed.
All arms use the hyperparameters selected in Section~\ref{sec:tuning}; none was
re-tuned for this table, which would have given the baselines a best-of-several
advantage that Omega-S did not receive.
Every arm in this table was measured in the same session, on the same hardware,
so that no comparison mixes the effect with the run-to-run variation quantified
in Section~\ref{sec:ruido}. The \emph{wins} column counts seeds on which
Omega-S keeps more code than that arm, on the absolute measure, and gives the
exact sign test and the Wilcoxon signed-rank test on the paired differences.}
\label{tab:main10}
\small
\begin{tabular}{lccccc}
\toprule
Arm & Code kept & Retention & Std & Omega-S wins & $p$ (signs / Wilcoxon) \\
\midrule
No regulariser & 0.173 & 62.9\% & 0.144 & 9/10 & 0.011 / 0.006 \\
Weight decay & 0.174 & 61.9\% & 0.090 & 10/10 & 0.001 / 0.002 \\
EWC ($\lambda{=}10^3$, swept) & 0.190 & 69.1\% & 0.185 & 8/10 & 0.055 / 0.014 \\
Row-norm control & 0.030 & 17.5\% & 0.093 & 10/10 & 0.001 / 0.002 \\
Cosine composite & 0.156 & 54.7\% & 0.125 & 10/10 & 0.001 / 0.002 \\
\textbf{Omega-S (log-ratio)} & \textbf{0.238} & \textbf{84.1\%} & 0.128 & --- & --- \\
\bottomrule
\end{tabular}
\end{table}

\begin{table}[ht]
\centering
\caption{\textbf{The two orientations of the modularity factor}, ten seeds
each, same hardware and same session. The framework defines $M$ as a modularity
measure and therefore places its inverse in the objective; the reference
implementation estimated algebraic connectivity, the inverse quantity. The
corrected orientation is the better of the two, but the paired difference
between them, $+0.075$, is smaller than the run-to-run standard deviation of
$0.104$ measured in Section~\ref{sec:ruido}, so \emph{the two orientations are
not distinguishable from each other} on ten seeds. What is distinguishable is
each against the baselines, and those comparisons are in
Table~\ref{tab:main10}.}
\label{tab:orient}
\small
\begin{tabular}{lccc}
\toprule
Orientation & Retention & Std & vs.\ the other \\
\midrule
As implemented ($M \approx \lambda_2$) & 76.6\% & 0.145 & 2/10 \\
\textbf{As defined ($M \approx 1/\lambda_2$)} & \textbf{84.1\%} & 0.128 & 8/10, $p=0.055$ \\
\bottomrule
\end{tabular}
\end{table}

(Table~\ref{tab:perseed10} is reported in the appendix.)

\begin{figure}[ht]
\centering
\includegraphics[width=0.82\textwidth]{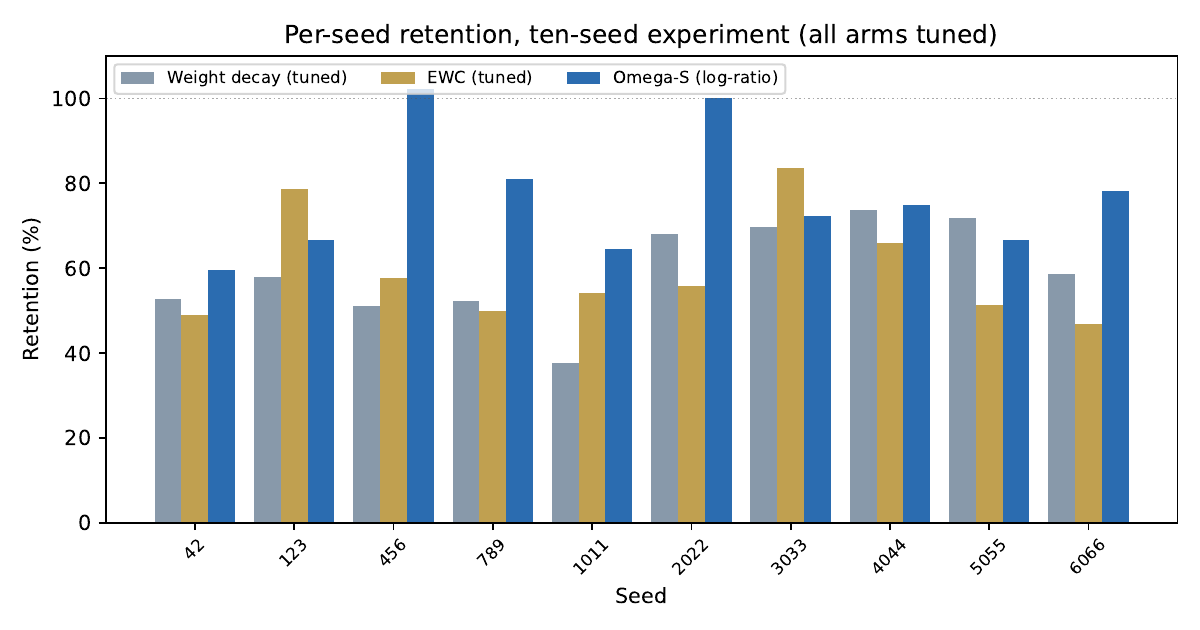}
\caption{Per-seed retention in the ten-seed experiment, all arms tuned.
Omega-S (right bars) exceeds both baselines on most seeds; the two seeds
where it trails EWC (123, 3033) are those where EWC itself is highest. The
dotted line marks retention $=100\%$ (full retention of the code capability
after the prose task).}
\label{fig:retention}
\end{figure}

In the ten-seed experiment, the log-ratio Omega-S improves retention over
\emph{tuned} weight decay on 10 of 10 seeds (mean $61.9\%\to84.1\%$,
$+17.3$\,pp) and over \emph{tuned} EWC on 8 of 10 ($+17.3$\,pp), and it does
so while ending with the highest absolute code capability of any arm
(HumanEval pass@1 $=0.238$ after the prose task, against $0.173$ with no
regulariser). We are careful about what that second number is. It is the
numerator of the retention ratio reported in absolute terms, and it is the
quantity a practitioner cares about, namely how much of the original
capability survives. It is \emph{not} a measure of plasticity, and we do not
describe it as one: the retention metric is HumanEval throughout while the
second task is prose, so nothing in this experiment measures performance on
the new task. Whether Omega-S costs plasticity on the second task is untested
here and would require a Wikitext perplexity measurement we did not make.

On the absolute measure Omega-S beats the no-regulariser arm on 9 of 10 seeds
(exact sign test, one-sided $p=0.011$), one more than the 8 of 10 the ratio
gives. The discrepancy is instructive. On seed 5055 Omega-S ends with more
code capability in absolute terms ($0.207$ against $0.171$) and yet a lower
retention ratio, because it had also reached a higher post-code baseline. The
ratio penalises an arm for doing better on the first task, which is the main
reason we lead with the absolute figure.

The penalty is applied during both tasks, so post-code HumanEval is
arm-specific by construction rather than a shared baseline; this is why we
lead with the absolute metric. The two are nonetheless statistically
indistinguishable between the Omega-S and no-regulariser arms: the means are
$0.291$ against $0.268$, Omega-S is ahead on 5 of 10 seeds with one tie
(exact sign test, one-sided $p = 0.50$), and the mean paired difference of
$+0.023$ sits well inside its own spread across seeds (sd $=0.050$). The
retention gain therefore does not arise from a different starting point.

We ran this experiment after a first five-seed run
(Table~\ref{tab:forgetting}) that showed the same direction more weakly
($4/5$ over weight decay); the effect strengthened, rather than regressing
to the mean, when we doubled the seeds and tuned every arm on the same
footing.

This comparison against EWC is worth stating in full, because Omega-S wins
it while asking for strictly less. EWC is the standard method for
catastrophic forgetting, and to build its penalty it must retain a copy of
the previous-task weights and estimate a Fisher information matrix by passing
\emph{previous-task data} back through the model, so it requires continued
access to the old data, which in production is often unavailable (privacy,
storage, data that has expired). Omega-S reads only the current weights: no
stored optimum, no Fisher matrix, no previous-task data. It therefore beats,
on $8$ of $10$ seeds, the field's standard forgetting baseline while removing
that baseline's core operational requirement. (We do not claim lower per-step
compute: Omega-S's penalty carries a Hutchinson estimate over $WW^\top$ whose
cost we have not benchmarked against EWC's; the advantage we do claim is
operational (no dependence on prior data), not raw speed.)

\paragraph{The margin is uneven, and its shape suggests a ceiling.} The
per-seed advantage is far from uniform: against the unregularised arm it ranges
from $-4.4$ to $+54.3$ percentage points. An earlier version of this work read
that spread as evidence of two regimes, topological on some weight graphs and
closer to norm-balancing on others, set by each seed's initialisation. We tested
that directly by measuring the four structural components on every seed
(Section~\ref{sec:mech}) and it is not supported: on all ten seeds the penalty
moves the same component, the variance of node degrees, and the other three do
not move at all. We record the two-regime hypothesis as refuted.

What the per-seed data do show is simpler. The gain over the unregularised model
shrinks as that model's own retention rises, with a rank correlation of
$\rho = -0.73$ across the ten seeds, and the single seed where Omega-S loses to
that baseline (42) is one where the baseline already retains two thirds of its
capability. The margin has the shape of a ceiling effect: the penalty helps
roughly in proportion to how much forgetting there is to prevent. We report this
as a pattern and not as a finding, because at $n = 10$ separating a genuine
ceiling from regression to the mean would need more seeds than we ran.

\paragraph{Not equivalent to norm-balancing.} A recurring concern for any
penalty on $WW^\top$ is that it may reduce, in practice, to equalising the norms
of the weight rows, which would make it a reparametrisation of weight decay
rather than a topological criterion. Table~\ref{tab:main10} includes the direct
control: a penalty on the variance of row norms, at the same calibration target,
run over the same ten seeds in the same session. It reaches 17.5\%
retention against 84.1\% for Omega-S, losing on all ten seeds
($p = 0.002$). Row-norm equalisation on its own does not buy retention in this
setting.

That control is run at Omega-S's calibration target rather than its own, which
is a deliberate but debatable choice: it tests whether the \emph{same} penalty
budget spent on norms achieves what spending it on degree variance does. A
version tuned on its own grid, and a protocol defect we found in the original
selection, are in Appendix~\ref{app:normbal}.

\subsection{What the penalty actually moves: a direct measurement}
\label{sec:mech}
\label{sec:mechanism}

Omega-S is \emph{by mathematical definition} a topological regulariser: its
objective is built from $\operatorname{Tr}(A^3)$, the third spectral moment
of the weight graph $A=WW^\top$, whose node-wise decomposition carries
information no purely spectral index determines (see Method).
That is a statement about the functional. It leaves open a separate,
empirical question: under the training conditions of these experiments,
\emph{which} of the objective's components does the penalty actually move to
produce the retention gain? We measured this directly rather than assuming
it.

For each of the ten seeds we trained task~A twice, once with no regulariser
and once with Omega-S, and recorded the four components of the objective
($C$ clustering, $D$ density, $M$ modularity, $\mathrm{Coex}$ degree
variance) on the effective LoRA weights, together with the retention gain of
that seed. The result is unambiguous and one-sided. The change in clustering
is $\Delta C \approx 0$ on every seed (e.g.\ seed~456: $C$ moves from
$0.999708$ to $0.999702$), whereas degree variance falls substantially
($\mathrm{Coex}$ drops $5$--$10\%$, e.g.\ seed~456: $201.6\to182.8$). In
other words, the retention effect is carried entirely by the denominator's
$\mathrm{Coex}$ term, not by the numerator's clustering term.

The reason is structural, not incidental. The pseudo-adjacency is built as
$A=\sigma(|WW^\top|)$, and the logistic $\sigma$ compresses all entries into
$[0.5,1)$; $A$ is therefore close to an all-ones matrix, on which the
normalised clustering $\operatorname{Tr}(A^3)/\lVert A\rVert_F^3$ is pinned
near its ceiling ($C\approx0.9997$) regardless of the underlying weight
structure. The clustering channel is saturated and inert \emph{by
construction}: Omega-S cannot raise $C$ because $\sigma$ has already removed
the contrast that would let it. What remains free to move is $\mathrm{Coex}$,
the variance of node degrees. We name this plainly as a design shortcoming
rather than a technical detail: the logistic map in the construction of $A$
is not motivated by the topological objective, and in retrospect it is
precisely what disables the clustering channel the method is built to
exploit. A construction that preserves contrast
(Appendix~\ref{app:reform}) would be preferable in any future formulation;
we report the saturation here, having measured it, rather than leaving it
for a reader to infer.

\paragraph{How much of the objective is actually live: a direct measurement of
all four factors.} The statement above concerns the clustering term, and a
reader is entitled to ask what the other three do. We measured the
\emph{elasticity} of each factor with respect to the weights, that is the
relative change in the factor produced by a relative change in $W$ along a
fixed random direction. The quantity is dimensionless, so it is comparable
across factors and modules, and a value of zero means the factor cannot
contribute gradient at all. Measured on the base weights of Llama-3-8B over ten
attention projections spanning layers 0 to 31, each with the branch the
implementation actually applies to it:

\begin{center}
\small
\begin{tabular}{lcccc}
\hline
& $C$ & $D$ & $M$ & $\mathrm{Coex}$ \\
\hline
Elasticity, base weights & $0.0000$ & $0.0000$ & $0.0001$ & $\mathbf{0.0091}$ \\
Elasticity, trained weights & $0.0000$ & $0.0000$ & $0.0001$ & $\mathbf{0.0139}$ \\
Relative change during a run & $-0.00\%$ & $+0.07\%$ & $+0.06\%$ & $\mathbf{-5.87\%}$ \\
\hline
\end{tabular}
\end{center}

\noindent The three rows measure three different things and agree. The first is
the sensitivity of each factor on the model's base weights; the second is the
same quantity after a full fine-tune, which tests whether a factor that looks
inert at initialisation comes alive once the weights move; the third is the
actual change in each factor over a run, relative to the unregularised arm,
median over the ten seeds.

Three of the four factors are numerically inert on all three measures, and one
carries everything. Degree variance responds roughly ninety times more strongly
than the modularity term, is the only factor above the noise floor, and is the
only one that moves during training, falling $5.87\%$ in median while the other
three stay under $0.1\%$. \textbf{As implemented and in these conditions, the
composite objective reduces in practice to a penalty on the variance of node
degrees.}

The middle row answers a question the first row alone leaves open. If a factor
looks inert at initialisation, it may still come alive once the weights move,
and the two orientations of $M$ do differ in outcome
(Section~\ref{sec:variantes}), which would be explained if $M$ started
contributing during training. It does not: the elasticity of $M$ is $0.0001$
before and after training alike, and its relative change over a run is
$+0.06\%$, indistinguishable from the clustering and density terms. Whatever
separates the two orientations, it is not the modularity term acting on the
weights, which is what leaves the operating-point account of
Section~\ref{sec:variantes} as the one that fits.

The result holds under both orientations of $M$: the numbers above are measured
with $M = 1/\lambda_2$, and the degree-variance change under the other
orientation was $-5$ to $-10\%$ with the other three factors equally inert. The
mechanism does not depend on which of the two is used, which is one more reason
to read the difference between them as a change of operating point rather than
as a topological effect.

It is worth being precise about what that inertia is a property of, because the
natural reading is that the framework's factors are weak and that is not what
the measurement says. The bounded map is what flattens them: in the ecological
setting from which the index derives, modularity is among the strongest signals
available, separating management types with $r^2 = 0.955$ in the survey of soil
communities that motivated the construction, comparable to the co-exclusion
term and higher than clustering. A factor that carries most of the signal in
the domain of origin and none under a saturating map applied to neural weights
is telling us about the map, not about the factor. The same holds for the
orientation question above: the direction of $M$ is fixed in the framework by
that ecological observation and not by convention, and it is only here, where
the term does not move, that the orientation ceases to matter.

We state the consequence plainly. \textbf{As currently formulated, and in
these conditions, Omega-S operates by reducing the variance of node degrees
in the weight graph}, a quantity that is strongly, though not perfectly,
correlated with the variance of weight-row norms (we measure
$r\approx0.60$ here, leaving roughly $40\%$ of degree-variance unexplained by
row norms). It is therefore not, in practice, a clustering-driven
topological method, even though its objective is topological in form. We
found no evidence of a second, clustering-driven regime: across all ten
seeds the movement is dominated by $\mathrm{Coex}$, so the bimodal per-seed
pattern of Table~\ref{tab:perseed10} is not explained by two structural
mechanisms in these data.

Three points follow, and we are careful not to overclaim in either
direction. First, this does \emph{not} weaken the empirical result: Omega-S
still leaves more code capability than no regularisation on $9/10$ seeds
($p=0.011$), and leads weight decay on $9/10$ and EWC on $8/10$ in retention,
without access to previous-task data. The method's demonstrated utility
stands; only the explanation of \emph{why} it works is revised, from
``clustering/topology'' to ``degree-variance control''. Second, degree-variance
control is itself a legitimate and, to our knowledge, uncharacterised lever
for LoRA continual learning, distinct from generic weight decay; the
row-norm control, now swept on its own grid, does not reliably beat no
regularisation ($6/10$) while Omega-S exceeds it on $8/10$, which bounds the
$\approx40\%$ unexplained component without fully quantifying it. Third, and this is a direction we offer to
the community rather than a result we claim here: the topological channel is
inert only because of the saturating $\sigma$. A reformulation that keeps
$C$ informative, for instance replacing $\sigma(|WW^\top|)$ with a contrast-
preserving map such as a temperature-scaled or rank-normalised affinity, or
penalising $\operatorname{Tr}(A^3)$ on a thresholded graph so that clustering
is not pinned near $1$, would let a genuinely clustering-driven variant of
Omega-S be tested. Whether such a variant outperforms the degree-variance
mechanism measured here is untested.
We sketch this reformulation in Appendix~\ref{app:reform} and leave its
evaluation to future work.

\paragraph{What the degree sequence measures, and a construction detail we
must declare.} Two further measurements refine the statement above. The
first is a discrepancy between the text and the reference implementation that
we report rather than leave for a reader to find. The implementation forms the
pseudo-adjacency as $\sigma(|WW^\top|)$ only when $W$ is non-square; when $W$
is square it applies $\sigma(|W|)$ elementwise, without forming the Gram
matrix. In Llama-3-8B, \texttt{q\_proj} is square and \texttt{v\_proj} is not,
so the runs reported here used one module of each kind and therefore two
constructions at once, while the description above and in the Method section
covers only the Gram form. The measured consequence is not cosmetic: on
\texttt{q\_proj} the degree-variance term differs by a factor of roughly $400$
between the two branches. This does not affect the reported numbers, which
were produced by the code as written, but it does affect how the method should
be described, and any reimplementation should fix the branch deliberately in
one direction or the other.

The second measurement asks what the degree sequence actually encodes, since
that is where the entire effect lives. Regressing the degree sequence of the
base weights on the row norms and partitioning the variance, the answer
differs by branch, and it differs in exactly the way the algebra predicts.
Where $A=\sigma(|W|)$ and the entries are small, the logistic is close to
linear, so the degree of a node is essentially the magnitude of its row and
column; empirically the row norms explain $96\%$ of the degree variance in
layers 8 to 31 of \texttt{q\_proj}. Where $A=\sigma(|WW^\top|)$, the degree is
the total absolute alignment of a row with all other rows; there the row norms
explain only $29\%$ on average across the same layers of \texttt{v\_proj}, and
the remaining $71\%$ is alignment structure that magnitude does not capture.
The regulariser is therefore applying two different criteria depending on
module shape: magnitude equalisation on square modules and alignment
equalisation on non-square ones. This is consistent with the row-norm control
being close to but not equal to Omega-S, and it suggests a specific ablation,
which we ran.

\paragraph{Which channel carries the effect: an ablation by module type.} We
restricted the penalty to one module type at a time while leaving the adapter
unchanged on both, so that what varies is which channel is regularised and not
where the model can learn. The comparison that matters is between the two
ablated arms rather than against the full objective: both concentrate the same
calibrated force on half as many modules, so they carry the same bias, whereas
comparing either to the full arm would confound channel with intensity.

\begin{table}[h]
\centering
\small
\begin{tabular}{lccc}
\hline
Seed & Only $v$ (alignment) & Only $q$ (magnitude) & Full objective \\
\hline
42 & 0.529 & 0.233 & 0.623 \\
123 & 0.775 & 0.385 & 0.878 \\
456 & 0.957 & 0.406 & 0.911 \\
789 & 0.619 & 1.000 & 1.056 \\
1011 & 1.000 & 0.296 & 0.708 \\
2022 & 0.633 & 0.366 & 0.825 \\
3033 & 0.800 & 0.342 & 0.837 \\
4044 & 0.723 & 0.645 & 0.855 \\
5055 & 0.621 & 0.800 & 0.741 \\
6066 & 0.837 & 0.515 & 0.976 \\
\hline
Mean & \textbf{0.750} & 0.499 & 0.841 \\
\hline
he\_A (task-A capability) & 0.289 & \textbf{0.193} & 0.287 \\
\hline
\end{tabular}
\caption{Retention with the penalty restricted to one module type, corrected
orientation, ten seeds. The adapter is unchanged in both arms; only the set of
modules the penalty acts on differs. \textbf{The last row is the reason this
comparison must be read with care}: the magnitude arm ends the first task with
$0.193$ HumanEval against $0.289$ for the alignment arm, so it learns task A
substantially worse and its retention ratio is measured against a lower
denominator. The direction of the difference is the same on the absolute
measure, but the magnitude of the gap is not interpretable.}
\label{tab:ablation}
\end{table}

Two things follow, and they point in different directions. \emph{The alignment
channel contributes more}: it leads on nine of ten seeds (exact sign test,
one-sided $p=0.011$; Wilcoxon $W=9$, two-sided $p=0.065$, which does not clear
the threshold because the single loss is also the largest difference), with a
mean paired advantage of $0.116$. \emph{But neither channel reproduces the full
objective}: the alignment arm exceeds it on three seeds of ten and the
magnitude arm on two, and the better of the two ($0.750$) exceeds the
no-regularisation baseline ($0.631$). The effect is not located in one channel;
it requires acting on both module types.

We also note an observation we cannot yet interpret. The single seed on which
the magnitude channel wins, 456, is the one where it wins by the largest margin
of any seed in the table, and where it \emph{exceeds the full objective}
($1.148$ against $1.023$). On seed 5055 the two channels tie and both exceed
the full objective. If this pattern held with more seeds it would suggest that
different initialisations fall into different regimes, one magnitude-driven and
one alignment-driven, which would be a candidate explanation for the bimodal
per-seed advantage of Table~\ref{tab:perseed10} that our earlier structural
measurement failed to find. On one seed of ten it is an observation and not a
finding, and we report it as such.

One caveat bounds the whole ablation. Because the calibration fixes the total
penalty gradient, restricting to half the modules concentrates that force on
half as many, so each ablated arm receives roughly twice the accumulated force
per module. This does not affect the comparison between the two ablated arms,
which carry the bias equally, but it does affect any comparison against the
full objective, and separating channel from intensity there would require a
further run at reduced target.

\subsection{How much run-to-run variation this setting carries}
\label{sec:ruido}

Every comparison in this paper is paired by seed, which presupposes that the
seed identifies the run. We tested that presupposition by repeating an identical
configuration: same seed, same code, same hardware, same arm. Four repetitions
of one arm and three of another gave standard deviations of $0.068$ and $0.104$
in retention ratio, with individual pairs as far apart as $0.596$ and $0.793$.
The same held when the four baseline arms were re-measured months after their
first run. Three of the four reproduced their means to within $0.04$; the
per-seed values did not, with a standard deviation of $0.104$ on the
differences. The fourth is worth naming because it changed a comparison: EWC,
whose mean rose from $0.593$ to $0.691$ on re-measurement. That is a shift of
$0.098$, the same order as the repeat-to-repeat deviation above, and it moved
Omega-S from winning 8 of 10 seeds against EWC on the retention ratio to
winning 6. Nothing about either method changed; the earlier EWC measurement
simply sat low in its own distribution. This is the concrete reason every arm
in Table~\ref{tab:main10} was measured in a single session, and it is what a
comparison against a previously published baseline figure risks.

Two consequences follow and we state both. First, the presupposition does not
hold in this setting: the seed fixes initialisation and data order but not the
non-deterministic reductions in the GPU kernels, so a paired test across seeds
is comparing runs that differ by more than the seed. Second, any mean difference
below roughly $0.066$, twice the standard error of a ten-seed mean, is not
distinguishable here. That bound is what places the two orientations of $M$ in
the not-distinguishable category while leaving the comparisons against the
baselines well outside it.

That GPU non-determinism dominates seed-to-seed variance is established
elsewhere. Morin and Willetts~\cite{morin2020} trained ResNet-50 fifty times
with a fixed seed and attribute $74$ to $87\%$ of the observed standard
deviation to it; Nagarajan et al.~\cite{nagarajan2018} report $12.4\%$ relative
deviation from the same source in reinforcement learning, and Pinto et
al.~\cite{pinto2021} find significant weight and prediction differences between
identically configured convolutional runs. We have not found the quantity
reported for low-rank fine-tuning of language models, which is the regime here,
and we report it because it bounds the reading of every seed-paired comparison
in this literature, ours included. The robust design would average several
repetitions per cell before pairing; we did not do that, and the cost of doing
it is a multiplication of the compute budget by the number of repetitions.

\subsection{One thing we would like others to run}
\label{sec:colab}

This paper closes with several things untested, listed in
Section~\ref{sec:lim}. Most need GPU hours and we do not expect anyone to spend
them on our behalf. One does not.

\textbf{The mechanism measurement of Section~\ref{sec:mech} runs on base weights
and takes minutes.} It reports the elasticity of each of the four factors with
respect to the weights, and it is what establishes that three of them are inert
under the construction used here:

\begin{center}
\texttt{python experiments/check\_M.py}
\end{center}

We have run it on one model. The claim that the composite objective reduces in
practice to a penalty on degree variance is a claim about \emph{that
construction on those weights}, and whether it holds elsewhere is the single
thing that would most change what this paper can say. \textbf{A report that the
factors are live on some other model is as useful to us as one that confirms
they are not}: it would mean the reduction is specific rather than general, and
we would rather know.

There is a reporting template in the repository.

\subsection{Limitations}

\label{sec:lim}
\label{sec:limits}

The primary limitation is scale: controlled structural experiments
(degree variance, FLOPs reduction) were conducted on a small MLP with
Split-MNIST. The main sequential fine-tuning result (mean retention
$61.9\%\to84.1\%$ over tuned weight decay, ten seeds, $10/10$ positive; $8/10$
over tuned EWC) is measured on Llama-3-8B with LoRA adapters on public
datasets using HumanEval as the retention metric, with the penalty verified
connected to the autograd graph. Three caveats bound the interpretation.
First, the per-seed advantage is bimodal (large wins, or losses only on the
two seeds where EWC itself peaks); ten seeds establish the direction but not
a per-seed structural model. Second, the mechanism is now measured
(Section~\ref{sec:mechanism}): the effect is carried by degree-variance
reduction, not clustering, so ``topological'' describes the objective, not
the operative mechanism in this formulation; how much of the degree-variance
effect is separable from row-norm equalisation is now measured rather than
assumed. A control swept on its own grid does not reliably beat no
regularisation ($6/10$) and Omega-S exceeds it on $8/10$, one-sided
$p=0.055$, which does not clear conventional significance at ten seeds. The
control's strength was moreover selected on two seeds that proved to be its
own worst two, so another strength might serve it better. On
GPT-2 (124M, full fine-tuning), a connected re-run finds that Omega-S
combined with group lasso does not exceed group lasso alone in structured
sparsity ($-0.72\%$ vs.\ $-0.77\%$ FLOPs). We report it because it bounds what
the method should be claimed to do: whatever it contributes, it is not
compression. The
FSDP environment used PCIe rather than NVLink; overhead ratios are valid but
should be re-confirmed in a higher-bandwidth environment.

A further limitation concerns what the name ``Omega-S'' covers. The
retention results use the log-ratio composite and the pruning results use the
raw trace penalty (Section~\ref{sec:twoforms}); these are different
objectives on different architectures and tasks, and the raw form is the arm
that performs worst on retention. Neither result is invalidated by this, since
each was produced by the objective it names, but the two should not be read as
evidence for a single method until the composite is evaluated on the pruning
setting as well.

Two further limitations follow from the work reported in
Section~\ref{sec:cosine}. The reference implementation builds the
pseudo-adjacency as $\sigma(|WW^\top|)$ only for non-square $W$ and as
$\sigma(|W|)$ for square $W$, so the runs reported here mix two constructions,
one per module type, while the Method section describes only the first. We
report this because it changes how the method should be described and because
the degree-variance term differs by roughly a factor of $400$ between the two
branches on \texttt{q\_proj}; the numbers themselves are unaffected, having
been produced by the code as written. And the negative result on the cosine
reformulation is specific to catastrophic forgetting under LoRA on this model
and task pair. It does not transfer to the index used as a monitoring
statistic in domains where the graph is built from correlations rather than
from weights, since there the clustering channel is not saturated to begin
with and the index is not used to intervene on the network.

% ============================================================
\section{Conclusion}

We have introduced Omega-S, a graph-based regulariser for large-scale
neural networks whose objective is built, via Hutchinson trace estimation,
from the clustering coefficient of the weight adjacency matrix $WW^\top$.
The method is computationally efficient ($\mathcal{O}(N^2)$ per layer),
adds negligible overhead to training ($+1.5$--$3.7\%$ at $K{=}10$), and
requires zero inter-GPU communication in LoRA+FSDP settings. In sequential
fine-tuning of Llama-3-8B, the log-ratio form of Omega-S leaves more code
capability than no regularisation at all on $9$ of $10$ seeds
($0.173\to0.238$ absolute HumanEval, $+37.7\%$ relative, $p=0.011$), and leads
weight decay on $9$ of $10$ and EWC on $8$ of $10$ in retention
($61.9\%\to84.1\%$), with zero access to previous-domain data. A first
five-seed run showed the same direction more weakly, so the effect
strengthened with sample size.

We are explicit about mechanism. Although the objective is topological by
construction, a direct measurement shows that in this formulation the
penalty acts through reduction of node-degree variance, not clustering: the
clustering channel is saturated by the logistic map used to build the
adjacency and does not move. Omega-S as it stands is thus best understood as
a degree-variance regulariser with a (currently inert) topological
objective; whether its benefit is separable from row-norm equalisation is now
tested rather than deferred, and a control swept on its own grid does not
reproduce the gain, though the margin over it does not clear conventional
significance at ten seeds. This revises the explanation
of the method, not its measured utility, which exceeds both standard
baselines. Restoring the topological channel through a contrast-preserving
construction was the obvious repair and we built it: it does revive the
clustering channel and it makes retention worse on all ten seeds
(Appendix~\ref{app:reform}). Whether some other construction would do better is
open, but the one that seemed most promising a priori does not.

A further limit concerns the scope of the regulariser. Weight decay,
used here as a retention baseline, is not primarily a forgetting method: it
is the standard tool for generalisation on a single task. Whether Omega-S
also competes with weight decay on that ground, that is, whether
degree-variance control improves single-task generalisation and not only
sequential retention, is a coherent follow-up we have not tested and flag
without prejudging the outcome.

Finally, Omega-S reflects a broader principle: structural regularisation
principles observed in biological systems can provide productive priors for
artificial neural network design. The formalisation of this principle, which runs from soil microbiome
network observations~\cite{ortiz2021,saati2026} to a mathematically
grounded and empirically validated regulariser for LLMs, represents the
central intellectual contribution of this work.

\paragraph{Patent.} The Omega-S method is the subject of a pending patent
application.

% ============================================================
\section*{Code Availability and Licensing}

The complete PyTorch implementation of Omega-S is publicly available at
\url{https://github.com/BiomeMakers/OmegaS-LLM}.

\textbf{Dual Licensing:} The source code is released under the
\textbf{GNU Affero General Public License v3 (AGPL-3.0)} for non-commercial
academic research, education, and non-profit experimentation. AGPL-3.0
permits free use, modification, and redistribution with attribution,
provided that any network deployment of a modified version includes public
disclosure of the modified source code.

\textbf{Commercial licensing}, including production deployment, integration
into commercial training pipelines or any use within a for-profit
organisation, requires a separate commercial licence that includes both
software and patent rights. Inquiries
regarding commercial licensing, research partnerships, or collaborative
development agreements should be directed to the author via the repository.

% ============================================================

% ============================================================
\appendix
\section*{Appendices}
\addcontentsline{toc}{section}{Appendices}

\noindent The material below is not needed to use the method or to judge the
results. It contains the framing that motivated the work, the related-work
survey, the experiments that support secondary claims, and the full record of
what we tried and it did not work. We keep it because the negative results are
part of what the paper reports, not because a reader needs them first.

\section*{Note on the appendices}

The main text reports the configuration this paper uses: the log-ratio
composite with $M = 1/\lambda_2$, with every arm measured in a single session.
Several appendices predate that configuration and report the alternative
orientation $M = \lambda_2$. They are: the calibration sweep that selected the
penalty strength; the \emph{strength sweep} of the row-norm control, which is a
separate experiment from the row-norm arm of Table~\ref{tab:main10} and asks
what that control achieves when tuned on its own grid; the hyperparameter
selection protocol for weight decay and EWC; and the cosine construction study.
We have not re-run them because what each establishes does not depend on the
orientation: the calibration sweep locates a target that both configurations
use, the row-norm sweep tests whether the effect reduces to norm balancing, and
the cosine study asks whether a construction that revives the clustering channel
helps, which Section~\ref{sec:mech} shows it does not under either orientation.

\textbf{The arms of Table~\ref{tab:main10} are not among them.} No regulariser,
weight decay, EWC, the row-norm control and the cosine composite were all
re-measured in one session with the configuration this paper uses, using the
hyperparameters those earlier sweeps had selected. What was not repeated is the
\emph{selection} of those hyperparameters, not their evaluation.

\textbf{Figures that differ between the two configurations are dated by
which appendix they appear in.} Where an appendix reports a mean retention of
$76.6\%$ for Omega-S, that is the alternative orientation; the configuration of
this paper gives $84.1\%$. We say this once here rather than repeating it in
each appendix.

\section{The row-norm control, swept on its own grid}
\label{app:normbal}

\noindent\emph{This appendix reports the alternative orientation
$M = \lambda_2$; see the note above. The row-norm arm of
Table~\ref{tab:main10} is separate and was measured with the configuration of
this paper.}

\paragraph{Not equivalent to norm-balancing: a control, and a protocol defect.}
A recurring concern for any penalty on $WW^\top$ is that it may reduce, in
practice, to equalising the norms of the weight rows, since degree and row norm
are strongly correlated. This is the objection that decides how novel the
mechanism is, so we tested it rather than argued about it.

Our first attempt at the test failed, and the failure was ours. The tuning
phase of the ten-seed run swept weight decay, EWC and the two Omega-S arms; it
did not sweep the row-norm control. In the evaluation phase that control was
launched with the calibration target selected for Omega-S, a value chosen to
make a \emph{different} penalty a fixed fraction of the task gradient. The
control was therefore the one arm never given a strength of its own, and it
collapsed (retention $20.7\%$, absolute HumanEval $0.027$, against $0.153$
already after the first task, so the model was damaged rather than merely
forgetful). Beating it established nothing. We found this by reading the
orchestration script rather than by reasoning about the numbers, and we report
it because an earlier version of this paper drew a conclusion from that arm
which the arm could not support.

We then ran the control properly, on a grid of its own, using the same two
selection seeds and the same ten evaluation seeds as every other arm. The grid
was placed below the Omega-S grid, since $0.03$ was already known to destroy
this penalty (Table~\ref{tab:rnsweep}). The selected target was $0.001$, at the
lower edge, which for this arm is the degenerate limit rather than an
unexplored optimum: as the target approaches zero the penalty vanishes and the
arm converges to no regularisation by construction.

\begin{table}[h]
\centering
\caption{Row-norm control, own hyperparameter sweep, on the same two selection
seeds (42, 123) used for every other arm. Retention rises monotonically as the
penalty weakens, which for this arm is convergence toward no regularisation
rather than an interior optimum.}
\label{tab:rnsweep}
\small
\begin{tabular}{lcc}
\toprule
Calibration target & Retention & Std \\
\midrule
\textbf{0.001} & \textbf{51.9\%} & 0.065 \\
0.003 & 39.6\% & 0.192 \\
0.01  & 31.5\% & 0.091 \\
0.03  & 28.3\% & 0.223 \\
\bottomrule
\end{tabular}
\end{table}

\begin{table}[h]
\centering
\caption{Per-seed absolute code capability (HumanEval pass@1 after the prose
task) for no regularisation, the row-norm control at its own selected strength,
and Omega-S. The last column is Omega-S minus the control.}
\label{tab:rnperseed}
\small
\begin{tabular}{lcccc}
\toprule
Seed & None & Row-norm & Omega-S & $\Delta$ \\
\midrule
42   & 0.134 & 0.146 & 0.152 & $+$0.006 \\
123  & 0.226 & 0.165 & 0.195 & $+$0.030 \\
456  & 0.165 & 0.171 & 0.274 & $+$0.103 \\
789  & 0.177 & 0.244 & 0.207 & $-$0.037 \\
1011 & 0.134 & 0.195 & 0.177 & $-$0.018 \\
2022 & 0.201 & 0.183 & 0.299 & $+$0.116 \\
3033 & 0.177 & 0.159 & 0.256 & $+$0.097 \\
4044 & 0.152 & 0.195 & 0.238 & $+$0.043 \\
5055 & 0.171 & 0.098 & 0.207 & $+$0.109 \\
6066 & 0.146 & 0.189 & 0.220 & $+$0.031 \\
\midrule
\textbf{Mean} & 0.168 & 0.174 & \textbf{0.223} & \textbf{$+$0.048} \\
\bottomrule
\end{tabular}
\end{table}

The result is a separation, and a qualified one. Tuned on its own grid, the
control does not reliably exceed no regularisation at all (6 of 10 seeds, mean
absolute $0.174$ against $0.168$, one-sided sign test $p=0.38$), and it carries
the highest variance of any arm in the study (retention standard deviation
$0.216$ against $0.145$ for Omega-S, with one seed collapsing to $0.254$).
Omega-S exceeds it on 8 of 10 seeds, one-sided $p=0.055$. Row-norm equalisation
on its own therefore does not buy retention in this setting, whereas Omega-S
does.

Two limits bound that reading and we state both. First, at ten seeds the margin
of Omega-S over the control does not clear conventional significance; it sits
at the same level as the margin over EWC. Second, the control's strength was
selected on seeds 42 and 123, which turned out to be its own two worst of the
ten ($0.453$ and $0.509$ retention, against $0.583$ to $0.914$ elsewhere), so
its selection is unreliable and another strength might serve it better. We did
not re-select on the evaluation seeds, because that would hand the control a
best-of-several advantage on the evaluation set which Omega-S never had.

\paragraph{The effective form is decisive.}
The raw $\operatorname{Tr}((WW^\top)^3)$ penalty applied directly to LoRA
factors (the form used in an earlier iteration of these experiments) is
the weakest arm in the same ten-seed sweep ($53.9\%$ mean retention, below
the $63.1\%$ no-regularisation baseline, winning only $1/10$ against weight
decay). The log-ratio \texttt{StochasticOmegaS} form is what delivers the
results above. We attribute the gap to scale sensitivity of the raw sextic
penalty, whose gradient vanishes as weights shrink; the log-ratio form is
approximately scale-invariant. Consequently, the library and the experiment
scripts are being unified on the log-ratio form, and earlier internal
figures that relied on the raw form are superseded.

\section{Background and Related Work}

\subsection{Topological Properties of Weight Graphs}

Neural network weight matrices can be interpreted as bipartite graphs where
neurons are nodes and weights define edge strengths. When the degree
distribution is highly skewed, with a small number of neurons capturing
most of the total norm, we term these high-degree nodes \emph{weight monopolies}.

The clustering coefficient, proportional to $\operatorname{Tr}(A^3)$,
quantifies the density of local connectivity clusters. We use it as a
structural descriptor and do not read a high value, on its own, as the
signature of a weight monopoly; as noted above, it is the degree variance that
distinguishes distributed from concentrated triadic structure.

\subsection{Scale-Free Networks in Biological Systems}

The scale-free network model~\cite{barabasi1999} describes graphs where
the degree distribution follows a power law $P(k) \sim k^{-\gamma}$.
Such networks exhibit high resilience to random node failures while
remaining vulnerable to targeted attacks on hubs.

Studies by Biome Makers and collaborators have characterised the topological
structure of microbial interaction networks across diverse agricultural
conditions. Ortiz-\'Alvarez et al.~\cite{ortiz2021} demonstrated that
farming practices in vineyard soils produce measurable disturbances in the
network properties of local fungal communities, with structural monopolies
emerging under high anthropogenic pressure. At global scale,
Saati-Santamar\'ia et al.~\cite{saati2026} showed that soil microbial
diversity and network organisation respond consistently to land use and
agricultural inputs worldwide, establishing degree distribution and hub
structure as reliable indicators of ecosystem health across biomes.
These topological properties are further formalised as diagnostic tools
in two Biome Makers patents: one on evaluating ecological disturbance via
network properties of organism communities~\cite{biome_patent}, and one that
defines resilience indices derived from the \emph{transitivity} of bacterial
and fungal co-occurrence networks~\cite{acedo2022indices}. The latter is the
direct antecedent of the present work. One qualification is needed for
accuracy: the \emph{global} transitivity of a graph is proportional to
$\operatorname{Tr}(A^3)$, whereas the vineyard study computes the average of
the \emph{local} transitivities, which captures the same triadic closure but is
a different functional and is not proportional to $\operatorname{Tr}(A^3)$. The
lineage runs through the notion of triadic closure, not through an identity of
statistics.

\subsection{Hutchinson Trace Estimator}

The Hutchinson estimator provides an unbiased stochastic approximation of
the trace of a matrix $M$:
\begin{equation}
  \operatorname{Tr}(M) \approx \frac{1}{n} \sum_{i=1}^{n}
  z_i^\top M z_i, \quad z_i \sim \text{Rademacher}(\pm 1).
\end{equation}
Applied to $M = A^3 = (WW^\top)^3$, the estimator avoids materialising
$A$ by decomposing the matrix-vector product:
\begin{equation}
  A^3 z = W\!\left(W^\top\!\left(W\!\left(W^\top\!\left(
  W(W^\top z)\right)\right)\right)\right).
\end{equation}
Each step requires $\mathcal{O}(mn)$ operations, yielding total cost
$\mathcal{O}(6 n_p m n)$ per layer versus $\mathcal{O}(m^3)$ for
explicit $A^3$. The Hutch++ variant~\cite{meyer2021} provides variance
reduction for ill-conditioned matrices.

\subsection{Related Regularisation Methods}

\textbf{Weight Decay / L2} pushes all weights uniformly toward zero,
agnostic to structural role. As we show empirically, it produces
unstructured sparsity not amenable to FLOPs reduction via structural
pruning.

\textbf{SAM}~\cite{foret2021} seeks flat minima in the loss landscape and
is blind to the internal topology of weight matrices. Omega-S and SAM are
complementary, not competing.

\textbf{Graph-spectral regularisers.} Fiedler
regularisation~\cite{tam2020} penalises the algebraic connectivity
$\lambda_2$ of the graph induced by $|W|$ during training, with theoretical
support from spectral graph theory and an equivalent formulation as a
structurally weighted $L_1$ penalty. It is the closest antecedent to the
present work in kind: a functional of a graph built from the weights,
minimised as part of the training objective rather than computed as a
diagnostic. Two things separate the two. First, the objective here is
triadic, built on $\operatorname{Tr}(A^3)$, and $\lambda_2$ appears only as
one of four factors; a graph-theoretic penalty defined on a bipartite
construction, as several in this family are, would annihilate the triadic
term identically. Second, and against our own interest, our measurement
(Section~\ref{sec:mechanism}) finds the $\lambda_2$ channel numerically
inert in this construction, so we make no claim to improve on Fiedler
regularisation on its own ground; the comparison remains to be run.

\textbf{Structured / Group Sparsity}~\cite{yuan2006} explicitly targets
neuron elimination based on weight magnitude. In our own measurements the
composite penalty combined with group lasso does not exceed group lasso alone
in the compression achieved, which is why we make no claim in that direction
here.

\subsection{Geometric and Spectral Approaches to Continual Learning}
\label{sec:cl_geometry}

A parallel and rapidly developing line of work attacks catastrophic
forgetting by reasoning about the \emph{geometry} of weight matrices rather
than the magnitude of individual parameters. Muon-OGD~\cite{lu2026muonogd}
argues that Frobenius-norm geometry is not the natural choice for
matrix-valued LLM parameters, and reformulates projection-based continual
learning under spectral-norm geometry, reporting gains on the standard and
15-task continual-learning benchmarks and on
TRACE~\cite{wang2023trace}. Related efforts constrain the spectral norm and
stable rank of routing manifolds in mixture-of-experts
architectures~\cite{srmoe2026}, or penalise functional drift within the
top-$k$ eigenspace of a task kernel~\cite{cflowrank2026}. Our premise, that the
topology of a weight matrix and not merely the size of its entries governs
how new learning overwrites old, is shared by this
literature and independently supported by it.

\textbf{A geometric law for forgetting, and a measurement of ours that speaks
to it.} Steele~\cite{steele2026} gives a geometric account of forgetting in
LoRA specifically, reporting that it follows
$\mathcal{F} = \alpha(1 - \cos^2\theta_{\min}) + \beta$, where
$\theta_{\min}$ is the minimum principal angle between the task gradient
subspaces, validated on synthetic tasks, Split-CIFAR100 with ViT-LoRA and
sequential GLUE with RoBERTa-LoRA. Two consequences of that law bear on the
present work.

The first is that we can report a measurement of the governing quantity on a
model of the scale evaluated here. In a separate screen we computed the
principal angles between the row spaces of two rank-8 LoRA deltas trained on
different tasks from a common Llama-3-8B backbone, across all sixty-four
attention projections. Against a simulated null for two random eight-dimensional
subspaces in $\mathbb{R}^{4096}$ (leading cosine $0.077$ on average, $0.100$ at
the maximum over three hundred draws), the leading principal cosine is $3$ to
$5$ times the null in fifty of the sixty-four modules, while the second and all
subsequent angles are indistinguishable from it. The overlap between two
adapters trained on different tasks is therefore essentially rank one: they
share one direction, and the rest of their agreement is chance. It also varies
systematically with depth, being largest in the first and last layers and
smallest in the middle, and five modules sit exactly at the null.

The second is a positioning point. Steele reports that orthogonalisation
methods such as O-LoRA add little when natural orthogonality is already high.
In our measurement seven of the eight directions are at the null, that is
already near-orthogonal, so on that reading an orthogonalisation approach would
have limited room in this setting. Omega-S does not orthogonalise; it acts on
the degree distribution of the weight graph, which is a different quantity, and
this is one reason the two families are complementary rather than competing.

\textbf{An objection from function space, which we record rather than
answer.} Hidekel and Raviv~\cite{cflowrank2026} argue in the NTK regime that
forgetting concentrates in a small number of old-task kernel eigenmodes, and
that this explains why regulariser acting in \emph{parameter} space can miss
interference in \emph{output} space. Omega-S is a parameter-space regulariser,
so the objection applies to it directly and we state it rather than passing
over it. Two things bound it in our setting without dissolving it. Their claim
is that parameter-space methods can miss such interference, not that they
cannot help: the retention results reported here are obtained by a
parameter-space penalty and hold on nine of ten seeds. And their account
predicts where a ceiling should lie, which is a testable statement about our
method that we have not tested. Measuring whether the residual forgetting under
Omega-S concentrates in the old-task kernel eigenmodes their theory identifies
would establish whether the two accounts describe the same limit, and we
identify it as an experiment we have not run.

\textbf{A sharper version of the same objection, and what separates our
method from its target.} Ning et al.~\cite{ning2026sae} make the empirical
case that weight-space regularisation underperforms on large language models
because the models are polysemantic: per-weight importance estimates of the
kind EWC computes are too coarse to isolate the knowledge that needs
protecting. They report that task-relevant representations are linearly
separable in a sparse-autoencoder feature basis but indistinguishable from
chance in the weight basis, and that weight-space protection is close to
non-selective at the concept level, and they regularise in activation space
instead. This is the most direct challenge to the family Omega-S belongs to
that we are aware of, and we would rather engage it than omit it.

The engagement is a distinction, not a rebuttal. Their evidence concerns
\emph{per-weight importance}: the claim is that no scalar attached to an
individual weight identifies what a concept occupies. Omega-S attaches no
importance to individual weights. It penalises a scalar property of the
\emph{whole matrix}, the variance of the degree sequence of the graph it
induces, which is a statement about how connectivity is distributed rather
than about which weights matter. Whether the polysemanticity argument extends
from per-weight importance to distributional properties of the weight matrix
is, as far as we can tell, open, and it is a question their framework is
better placed to settle than ours. We note only that the retention results
reported here were obtained in the weight basis, which the argument as stated
would not predict.

\textbf{A comparison of magnitude worth recording.} Independently,
Laitinen-Fredriksson Lundstrom-Imanov~\cite{lundstrom2026} report a
mechanistic study of forgetting across twenty models and find that a
curvature-regularisation intervention reduces forgetting by $34\%$ with
minimal impact on new-task learning, a figure close to the $33\%$ relative
gain in retained capability reported here. The two are not commensurable, the
settings, models and metrics differing throughout, and we place them side by
side only to indicate that interventions acting on the geometry of the loss
or of the weights appear to reach effects of a similar order, which is
information a reader deciding whether this line is worth pursuing should
have.

\textbf{Two method classes, and where Omega-S sits.} It is useful to
separate two families. \emph{Penalty} methods add a term to the loss that
discourages changes deemed harmful: EWC~\cite{kirkpatrick2017} weights those
changes by Fisher information, LwF~\cite{li2017lwf} by output distillation.
\emph{Projection} methods instead modify the update direction itself,
restricting it to a subspace orthogonal to protected past-task directions:
OGD~\cite{farajtabar2020}, O-LoRA~\cite{wang2023olora}, OSFT and Sculpting
Subspaces~\cite{nayak2025}, and Muon-OGD. On the multi-task benchmarks
reported in~\cite{lu2026muonogd}, projection methods substantially
outperform penalty methods: 75.8 (O-LoRA) and 78.9 (Muon-OGD) against 50.3
(EWC) and 52.3 (LwF) average accuracy on the standard benchmark. Omega-S belongs to the penalty family, and we state
this explicitly rather than leaving it to be inferred.

\textbf{Scope of the comparison we do and do not make.} We do not report a
head-to-head comparison against these methods, and we are precise about
why. The setting evaluated here is single-adapter LoRA fine-tuning with
retention measured on a held-out capability after a task pair
(Section~\ref{sec:stability}), whereas the works above evaluate multi-task
streams of five to fifteen tasks under full or multi-adapter fine-tuning.
The numbers are therefore not commensurable, and quoting them side by side
would be misleading. What follows from this is an obligation rather than a
defence: extending Omega-S to those benchmarks, with EWC as the
class-appropriate baseline and O-LoRA or OSFT as the strong incumbents, is
the natural next validation step, and it is specified as
\emph{Experiment~A} in the distributed validation protocol accompanying
this work.

\textbf{An open problem where Omega-S may be directly relevant.}
Projection methods must decide \emph{which} directions to protect.
Current practice, including Muon-OGD's low-rank instantiation, takes the
top-$k$ left and right singular vectors of the pretrained weight matrix,
and~\cite{lu2026muonogd} names identifying these constraint directions more
effectively as explicit future work. The per-node
decomposition $\mathrm{diag}(A^3)$ used by Omega-S ranks units by a
different criterion: the concentration of closed three-step walks incident
on each unit, which is a function of the eigenvectors as well as the
eigenvalues, and is therefore not recoverable from the singular-value
spectrum alone. In preliminary probing on TinyLlama-1.1B we find this
ranking to be largely uncorrelated with Fisher-information importance
(mean Spearman $\rho \approx 0.02$ across 88 attention projections, after
controlling for row norm), indicating that it carries information distinct
from the standard parameter-importance signal. Whether it also carries
information distinct from the dominant singular directions, and whether
substituting or augmenting the protected subspace with it improves
retention, is untested. We flag it as a concrete point of contact with this
literature and an open invitation to collaboration rather than as a result.

\subsection{Non-Uniform Structured Pruning of LLMs}

A growing body of work demonstrates that uniform sparsity across transformer
layers is suboptimal~\cite{yin2024owl}, motivating \emph{non-uniform}
sparsity allocation strategies that assign differentiated pruning budgets
per layer.

\textbf{OWL}~\cite{yin2024owl} allocates higher sparsity to layers with
fewer activation outliers, mapping the outlier distribution linearly to
sparsity ratios. It demonstrates consistent improvements over uniform Wanda
and SparseGPT at high sparsity levels.

\textbf{AlphaPruning}~\cite{lu2024alpha} estimates per-layer importance
using Random Matrix Theory, specifically the heavy-tailed spectral
exponent of the weight matrix. Layers with heavier-tailed spectra are
considered more important and receive lower sparsity. This is the closest
prior work to Omega-S in spirit: both use a structural property of the
weight matrix (rather than activation statistics) to guide pruning.

\textbf{SV-NUP}~\cite{svnup2025} quantifies per-layer contribution via
Shapley values, enabling tailored pruning budgets that preserve the most
critical layers. Notably, \cite{svnup2025} also cites topological insights
into restricted Boltzmann machines~\cite{mocanu2016} as motivation, the
same class of structural observations that underlies Omega-S.

\textbf{DarwinLM}~\cite{tang2025} uses evolutionary search to discover
optimal non-uniform sparsity allocations, trading off computation for
accuracy.

\textbf{Distinction from Omega-S.} These methods share with Omega-S the
premise that weight-matrix structure carries actionable information, but
they differ in \emph{stage}, \emph{objective}, and \emph{quantity}. OWL and
AlphaPruning operate \emph{after} training as allocators of a compression
budget; their target is inference efficiency at fixed accuracy. Omega-S
instead acts \emph{during} training as a regulariser on the topology of each
weight matrix, and its target is \emph{knowledge retention}, that is, resistance to
catastrophic forgetting under sequential fine-tuning, rather than
compression. Where AlphaPruning \emph{measures} the heavy-tailedness of a
layer's spectrum to decide how much to prune, Omega-S \emph{reshapes} the
weight-graph structure during optimisation to reduce its concentration
(degree variance). The two directions are therefore complementary rather
than competing: a model regularised with Omega-S can still be pruned
afterward by OWL or AlphaPruning, and the questions they answer, namely
\emph{how much of each layer to remove} versus \emph{how to train weights
that forget less}, do not overlap. We do not claim Omega-S as a pruning-allocation
signal; its contribution is at training time, on retention.

% ============================================================
\subsection{Conceptual Origin: From Soil Microbiomes to Neural Networks}

The conceptual origin of Omega-S lies not in the loss landscape geometry
of deep learning, but in the observed topological properties of biological
information networks, specifically the connectivity structure of microbial
communities in agricultural soil ecosystems.

Empirical studies of soil microbiome networks report that agricultural
disturbance systematically reorganises the association structure of the
community. In a survey of 350 vineyard soils across two
countries~\cite{ortiz2021}, soils under low-intervention management showed a
higher clustering coefficient, lower modularity and a lower proportion of
co-exclusion edges than conventionally managed soils, a combination the
authors read as denser, less niche-partitioned communities with greater
resistance to species loss. In a global survey of 1{,}921 soils from 33
countries~\cite{saati2026}, agricultural inputs consistently reduced network
density across both bacterial and fungal communities, which the authors
associate with reduced functional redundancy.

We are deliberate about what we do and do not take from this. The two studies
use different association estimators, different units of analysis (one network
per sample against one network per condition) and different clustering
definitions, so their individual factors are not directly comparable and we do
not pool them into a single directional claim. Neither reports degree
distributions or power-law fits, so we make no scale-free claim here; an
earlier version of this paper did, and it was not supported by the sources
cited. What the two share is that disturbance reorganises network structure in
directions their authors link to lower resilience. The intuition we carry
across is accordingly a weak one: how connectivity is \emph{distributed} over a
network, and not only how much of it there is, carries information about
robustness that node counts and diversity indices do not.

This biological observation raises a compelling question for artificial neural
networks: do similar structural monopolies emerge during training, and if so,
do they carry analogous costs in terms of robustness, generalisation, and
efficiency?

The answer, we argue, is yes. During the training of large language models,
a small subset of neurons progressively concentrates disproportionate weight
connectivity across layers, creating highly non-uniform degree distributions
in the implicit weight graph. This concentration, which we term
\emph{weight monopolies}, persists even under strong weight decay and
sharpness-aware optimisation, which target the loss landscape geometry but
leave the internal topology of weight matrices untouched.

\subsection{Thermodynamic and Information-Theoretic Framing}

The connection between biological network topology and artificial neural
networks has a deeper thermodynamic grounding. In physical systems, the
principle of minimum energy dissipation (related to Boltzmann's entropy
formulation) predicts that stable, efficient systems minimise redundant
energy flows by distributing load across available channels. Structural
monopolies represent thermodynamically inefficient configurations: they
concentrate functional load on a subset of nodes while leaving the remaining
capacity idle.

The clustering coefficient of a network, quantified as the normalised
count of closed triangles in the connectivity graph and proportional to
$\operatorname{Tr}(A^3)$, is a sensitive measure of how triadic structure is
distributed. We deliberately do not attach a direction to it. The same
triangle count is compatible with connectivity spread evenly across the graph
and with connectivity concentrated on a hub and its core, and what separates
those two configurations is the variance of the node degrees rather than the
triangle count itself (Section~\ref{sec:mechanism} and~\cite{acedo2026fsri}).
What we retain here is only that $\operatorname{Tr}(A^3)$ is a structural
quantity distinct from parameter magnitude, and that its gradient is cheap
enough to apply during training.

This framing connects Omega-S to a broader principle: just as the curvature
of spacetime governs the trajectories of physical systems in general
relativity, the topological curvature of the weight connectivity graph
governs the flow of information through a neural network. Standard
regularisers operate on the metric of the loss landscape; Omega-S operates
on the metric of the network's internal topology.

\subsection{Relation to Spectral Quantities}
\label{sec:spectral_relation}

We state a fact about $\operatorname{Tr}(A^3)$ that a reader may otherwise
raise as an objection, and explain why it does not undermine the use made of
it here. For symmetric $A$ with eigenvalues $\{\lambda_i\}$,
\begin{equation}
\operatorname{Tr}(A^3) = \sum_i \lambda_i^3 ,
\label{eq:third_moment_llm}
\end{equation}
so the quantity is exactly the third moment of the spectrum of $WW^\top$. It
is therefore not an alternative to spectral analysis but a particular
functional of the spectrum, in the same family as the effective rank, the
Vendi score and the Absorption Ratio. As a \emph{descriptive statistic} it
carries no information those measures lack, and we make no claim that it
does; the companion theoretical work develops this point and its
consequences~\cite{acedo2026fsri}.

The relevant observation is that Omega-S is not used here as a statistic. It
is used as a \emph{training objective}, and what an objective contributes is
its gradient, not its value. Different functionals of the same spectrum have
different gradients and therefore induce different geometries:
\begin{equation}
\nabla_W \operatorname{Tr}\!\left((WW^\top)^3\right) = 6\,(WW^\top)^2 W ,
\end{equation}
a polynomial expression evaluable through matrix--vector products alone. The
entropy-based functionals require the eigenvalues themselves, so
differentiating them entails an eigendecomposition at every optimisation
step, which is expensive at LLM scale and numerically delicate where
eigenvalues are near-degenerate. Combined with the Hutchinson estimator, which avoids
materialising $A$ at all, this is what makes the penalty tractable as a
per-step term rather than a diagnostic computed occasionally. The
contribution claimed here is accordingly narrow and specific: not that
penalising spectral structure is novel, but that \emph{this} functional, with
\emph{this} estimator, is cheap enough to apply during training and improves
retention when so applied.

A second consequence of Eq.~\eqref{eq:third_moment_llm} concerns where the
non-redundant information in the framework resides. The global scalar is
spectrally determined; the per-node decomposition $\mathrm{diag}(A^3)$ is
not, since $(A^3)_{ii} = \sum_k \lambda_k^3 (v_k)_i^2$ depends on the
eigenvectors. This is the basis of the attribution discussed in
Section~\ref{sec:cl_geometry}.

\subsection{Structural Control: Omega-S vs.\ Weight Decay}

\begin{table}[ht]
\centering
\caption{Structural control metrics across training configurations.}
\label{tab:structural}
\begin{tabular}{lcccc}
\toprule
Architecture & Eval Loss ($\downarrow$) & Accuracy ($\uparrow$) &
Degree Var.\ ($\downarrow$) & Max Hub ($\downarrow$) \\
\midrule
Baseline             & 0.1650 & 95.46\% & 41.94  & 125.47 \\
Omega-S              & 0.1894 & 94.92\% & \textbf{0.136} & 116.83 \\
Weight Decay ($\lambda{=}0.01$) & 0.2464 & 92.56\% & 1.12 & \textbf{69.00} \\
\bottomrule
\end{tabular}
\end{table}

Omega-S achieves 8$\times$ greater reduction in degree variance compared
to weight decay (0.136 vs.\ 1.12), with only 0.54\,pp accuracy cost
versus 2.90\,pp for weight decay. Weight decay reduces the maximum hub
more aggressively (69.00 vs.\ 116.83), consistent with indiscriminate norm
suppression. We interpret this as evidence that the two methods target
different aspects of weight concentration: Omega-S redistributes the global
degree distribution while preserving hubs that serve legitimate routing
functions, the neural analogue of the microbial network observation that
not all hubs are monopolies.

\subsection{How the hyperparameters were selected, and the anomaly this explains}
\label{sec:tuning}

Table~\ref{tab:main10} contains a result a reader will notice, and we explain it
rather than leave it standing: both regularised baselines retain \emph{less}
than the arm with no regulariser at all ($59.3\%$ and $59.3\%$ against
$63.1\%$). The explanation is the selection protocol, which is weaker than an
earlier version of this paper described.

Every arm was selected, and we give the grids. Weight decay was swept over
$\{0,\,0.01,\,0.05,\,0.1\}$; the grid included zero, so the procedure was free
to switch the regulariser off, and it did not, returning $\lambda = 0.05$.
EWC's $\lambda$ was swept over $\{10^2,\,5{\times}10^2,\,10^3,\,5{\times}10^3,\,
10^4\}$ and returned $10^3$, which was both the highest-scoring value and by
some margin the most stable across the selection seeds (standard deviation
$0.011$, against $0.06$ to $0.08$ for the others). That value coincides with
the library default, so the sweep confirmed the default rather than changing
it; a concern recorded at the time, that EWC might be running at a badly set
$\lambda$, was checked and dismissed rather than left standing. Omega-S's
calibration target was swept over $\{0.03,\,0.1,\,0.3,\,0.5\}$ and returned
$0.03$.

The limitation is therefore not that the arms went unselected. It is that the
selection was made on \emph{two} seeds, and both of them are among the ten
later used for evaluation, so the selection set is a subset of the evaluation
set rather than disjoint from it. The retention observed during selection
($0.69$ for weight decay, $0.72$ for EWC) is well above what the same arms
attain over the full ten seeds ($0.59$ and $0.59$), which is the regression one
expects when a small and favourable selection set is expanded, and it is why
both tuned arms can sit below the no-regulariser arm without either method
being at fault.

Three consequences follow, and we keep them separate.

\emph{First}, the two baselines in Table~\ref{tab:main10} should be read as one
reasonable fixed setting of each method, not as the best each method can do. A
properly selected weight decay would plausibly land at or above the
no-regulariser baseline, and the $+17.3$\,pp margin over it should not be quoted
as though it were.

\emph{Second}, the comparison that does not depend on this protocol is the one
against the no-regulariser arm, which has no hyperparameter to select and is
therefore unbiased on these ten seeds. Omega-S beats it on \textbf{8 of 10
seeds}, mean $63.1\% \to 76.6\%$ ($+13.5$\,pp; exact sign test, one-sided
$p = 0.055$, two-sided $p = 0.11$). We regard this as the conservative headline
and prefer it to the margin over weight decay. For completeness the same test
gives $9/10$ against weight decay (one-sided $p = 0.011$) and $8/10$ against EWC
(one-sided $p = 0.055$).

\emph{Third}, Omega-S's own selection carries the same weakness, and one of its
own: it was selected on the same two seeds, and the value returned, $0.03$,
sits at the edge of the swept range, with retention falling monotonically as
the target rises. A grid whose optimum lies on its boundary has not been shown
to contain the optimum. Running the selection on seeds disjoint from the
evaluation set, and extending the Omega-S grid below its current floor, are the
first two things we would change, and where a replication should differ from
us.

\subsection{Testing the contrast-preserving reformulation: a negative result}
\label{app:cosine}
\label{sec:cosine}

The mechanism measurement above left one question unresolved: the clustering
channel is inert only because of the saturating
$\sigma$, so a construction that preserves contrast would let a genuinely
clustering-driven variant be tested, and whether such a variant outperforms
the degree-variance mechanism is unknown. We subsequently built that variant
and tested it. This section reports the outcome, which is negative and, we
think, informative: activating the clustering channel does not improve
retention, it degrades it, and it does so monotonically.

\paragraph{Which constructions we screened, and why this one.} Restoring an
informative clustering channel means replacing the bounded map that destroyed
it, and several candidates are available. We screened six on the base weights
before committing to any of them, using the excess statistic $C/D$, which
equals exactly $1$ for a constant affinity matrix and exceeds $1$ in the
presence of triadic structure. The screen is cheap, runs on a laptop, and
falsifies most of the candidates outright.

\begin{table}[h]
\centering
\small
\begin{tabular}{llc}
\hline
Construction & Form & Excess $C/D$ \\
\hline
Original (reference) & $\sigma(|WW^\top|)$ & $1.0000$ \\
Original, zero diagonal & as above, $A_{ii}=0$ & $1.0000$ \\
Temperature-scaled & $\sigma(|WW^\top|/\tau - b)$ & $1.0000$ for all $\tau\in[0.1,50]$ \\
Degree-normalised & $D^{-1/2}AD^{-1/2}$ & $0.9997$ (spread $0.0035$) \\
Rank-normalised & empirical CDF of $|WW^\top|$ & decompresses ($C\!:\,1.0\to0.69$) \\
\textbf{Row-wise cosine} & $|\cos(w_i,w_j)|$, $A_{ii}=0$ & $\mathbf{1.357}$ median, 28 modules \\
Doubly stochastic & Sinkhorn scaling of $|WW^\top|$ & $1$ by construction \\
\hline
\end{tabular}
\caption{Screening of contrast-preserving constructions on base Llama-3-8B
weights. Zeroing the diagonal does not help, which rules out diagonal
dominance as the cause. Temperature scaling fails at every temperature
because $|WW^\top|$ spans many orders of magnitude, so dividing by $\tau$ does
not prevent saturation. Degree normalisation, the option with the strongest
prior support in the clusterability literature, also fails: normalising a
matrix the logistic has already flattened does not restore contrast, because
the problem is one of scale upstream. Two candidates survive, and the cosine
form was chosen over the rank form because the rank map is piecewise constant
and therefore gives a noisy or vanishing gradient. The doubly stochastic form
is listed for completeness: it nulls the excess exactly, which makes it useful
as an attribution control rather than as a training construction.}
\label{tab:constructions}
\end{table}

Two further observations from the same screen bear on the choice. First,
elasticity: under the original construction the median elasticity of
$\operatorname{Tr}(A^3)$ with respect to a perturbation of $W$ is $0.84$, and
in at least one module it is exactly zero in float32, meaning the penalty
cannot move the objective there at all. Under the cosine form the median is
$16.6$, roughly twenty times more responsive. Second, the signal is not
uniform across module types: on \texttt{down\_proj} the cosine excess is
$1.3\%$ at the median against $36\%$ in attention, and it decays monotonically
with depth ($11.3\%$ at layer~0, $0.33\%$ at layer~31), which is why the
experiments that follow apply the construction to the attention projections.

\paragraph{Construction.} Of the six candidate constructions screened above,
the one that both removes the saturation and admits a clean algebraic argument
is the row-wise cosine affinity
\begin{equation}
A^{\cos}_{ij} = \bigl|\cos(w_i, w_j)\bigr|
= \frac{|\langle w_i, w_j\rangle|}{\lVert w_i\rVert\,\lVert w_j\rVert},
\qquad A^{\cos}_{ii} = 0 .
\label{eq:cosine}
\end{equation}
Two properties motivate it. First, it is invariant to row scaling by
construction: normalising a row changes no cosine, so the node degree
$k_i$ becomes mathematically independent of $\lVert w_i\rVert$. This is the
property the original construction lacks, where $k_i$ contains
$\lVert w_i\rVert^2$ through the diagonal of $WW^\top$. Second, the entries
occupy the full $[0,1]$ range rather than collapsing to the midpoint of a
bounded map.

\paragraph{Static screening.} Before spending compute we verified on the base
weights of Llama-3-8B that the construction actually decompresses the
clustering channel. We use the excess statistic $C/D$, which equals $1$
exactly for a constant affinity matrix and exceeds $1$ in the presence of
triadic structure. Under the original construction $C/D = 1.0000$ to four
decimal places in every module examined, and in at least one module the
elasticity of $\operatorname{Tr}(A^3)$ with respect to a perturbation of $W$
is exactly zero in float32: the channel is not merely weak, it is
numerically dead. Under Eq.~\ref{eq:cosine} the same statistic gives a median
excess of $1.36$ across 28 modules (per-module range $1.05$ to $1.84$), and
the elasticity is roughly twenty times larger. Measured with the library's own
normalisation $\operatorname{Tr}(A^3)/\lVert A\rVert_F^3$, which is the
quantity that actually enters the objective, the clustering term moves from
$0.9999$ to $0.802$ in \texttt{q\_proj} and from $0.99999$ to $0.512$ in
\texttt{v\_proj}. The channel is live.

\paragraph{The channel is controllable under training.} Static structure need
not survive optimisation, so we screened that separately before evaluating
retention. Training with a clustering-only penalty
$\pm\lambda\log\operatorname{Tr}(A^3_{\cos})$ and diagnosing the exact excess
statistic every 25 steps, the sign of the penalty separates the direction of
movement in every module and at every dose tested: a positive coefficient
lowers $C/D$ and a negative one raises it, in 16 of 16 measurements across
\texttt{q\_proj}, \texttt{k\_proj}, \texttt{v\_proj} and \texttt{o\_proj}. The
between-sign contrast grows monotonically with penalty strength. The
clustering channel is therefore not only live in the weights, it is
steerable during LoRA training, which the original construction is not.

\paragraph{Recalibration.} The ratio calibration fixes the penalty gradient at
a target fraction of the task gradient, so it normalises magnitude but not
direction. Because the cosine construction changes the direction of the
penalty gradient, the optimal target does not transfer. Swept on two seeds
held out of the evaluation set, the optimum is interior and identical in rank
order on both seeds ($0.003 > 0.01 > 0.03 > 0.001$), an order of magnitude
weaker than the optimum of the original construction. We note that the
original sweep selected its target at the edge of its grid; extending the grid
downwards here is what allowed the optimum to be bracketed on both sides.

\paragraph{Retention outcome.} We then evaluated the full composite objective
with the cosine construction, changing nothing else: the same four factors,
the same formulas, the same modules, the same ten evaluation seeds. Against
the original formulation it loses on \textbf{all ten seeds} (mean retention
$0.537$ against $0.766$; mean paired difference $-0.229$; Wilcoxon signed-rank
$W=0$, two-sided $p=0.002$). Every individual value of the cosine arm falls
below every individual value of the original arm, so the comparison does not
depend on the pairing. Against weight decay and EWC the cosine arm is a
statistical tie with the point estimate below ($4/10$ each, mean difference
$-0.056$), and it also falls below the no-regularisation arm.

\paragraph{Decomposition.} Running the clustering-only arm on the same ten
seeds at the same operating point isolates the contribution of the remaining
three factors. Their median contribution is $+0.147$ under the cosine
construction against $+0.227$ under the original one, that is roughly
two-thirds as much, while the clustering-only arm itself drops from $0.539$
(original construction, where the term is inert) to $0.477$ (cosine, where it
is active). The arithmetic is consistent: the loss is accounted for by the
clustering term changing from inert to actively harmful, plus a smaller
reduction in what the other factors contribute. We note that the
clustering-only arm has by far the highest between-seed variance of any arm we
have run (standard deviation $0.35$, range $0.09$ to $1.33$), so the pure
clustering channel is capable of large gains on individual seeds but not
reliably.

\paragraph{What we take from this.} The saturated clustering channel was not a
defect holding the method back. Across the three arms the ordering is
monotone in how active the clustering term is: clustering-only under cosine
($0.477$), full composite under cosine ($0.537$), full composite under the
original construction ($0.766$). The more the clustering channel moves, the
worse the retention. The honest reading is that the original formulation works
\emph{because} it is, in effect, a degree-variance regulariser, and that adding
a live topological channel degrades it rather than improving it. This closes
the open direction stated in the mechanism section and in
Appendix~\ref{app:reform}, in the negative.

We flag two limits on this conclusion. It is specific to catastrophic
forgetting under LoRA on this model and task pair, and says nothing about the
index used as a monitoring statistic in domains where the graph is built from
correlations rather than weights. And the cosine construction is applied
uniformly to all modules, whereas the original implementation branches on
matrix shape (see Section~\ref{sec:limits}), so for square modules the
comparison differs in two respects rather than one.

\subsection{First run: five-seed replication}
\label{sec:fiveseed}

Before the ten-seed experiment we ran a smaller five-seed version with the
same protocol but with Omega-S at its \emph{calibrated} strength (penalty
set to a fixed fraction of the task gradient) rather than the swept target
used above. We report it as a separate experiment because the Omega-S
configuration differs; the two should be read as independent runs, not
pooled.

\begin{table}[ht]
\centering
\caption{First run, five seeds, calibrated Omega-S. Retention as in
Table~\ref{tab:main10}. The direction matches the ten-seed result but is
weaker and, against EWC, not separable at this seed count.}
\label{tab:forgetting}
\begin{tabular}{lcccc}
\toprule
Seed & Weight decay & EWC & Omega-S & $\Delta$ vs WD \\
     & (tuned)      & ($\lambda{=}10^3$) & & \\
\midrule
42   & 48.94\% & 51.22\% & 50.00\% & $+$1.1\,pp \\
123  & 72.34\% & 77.78\% & 72.22\% & $-$0.1\,pp \\
456  & 56.10\% & 66.67\% & 66.15\% & $+$10.1\,pp \\
789  & 60.47\% & 65.91\% & 92.86\% & $+$32.4\,pp \\
1011 & 50.00\% & 39.13\% & 77.08\% & $+$27.1\,pp \\
\midrule
\textbf{Mean} & 57.57\% & 60.14\% & \textbf{71.66\%} & \textbf{$+$14.1\,pp} \\
\bottomrule
\end{tabular}
\end{table}

In this first run Omega-S beat tuned weight decay on 4 of 5 seeds
($+14.1$\,pp mean). Against EWC the five seeds did not separate the two
methods: Omega-S had the higher mean ($71.7\%$ vs.\ $60.1\%$) but EWC
retained more on three of the five, with per-seed spread comparable to the
gap. The ten-seed experiment, with Omega-S tuned rather than calibrated,
is what resolves this: there Omega-S leads EWC on $8/10$. Both runs agree on
the robust conclusion (Omega-S over weight decay), and the larger run
additionally separates it from EWC. We keep both tables so the reader can
see the effect grow with sample size rather than shrink.

In both runs Omega-S ends with the highest absolute post-prose HumanEval of
all arms ($0.223$ at ten seeds against $0.168$ with no regulariser and $0.162$
for weight decay). We report this as retained capability, not as plasticity,
for the reason given in Section~\ref{sec:stability}. All of this is obtained with zero
architectural modification, zero access to previous domain data, and a
training overhead of $+1.5\%$ (Section~\ref{sec:infra}).

\subsection{Infrastructure Overhead: Single-GPU and FSDP}
\label{sec:infra}

\begin{table}[ht]
\centering
\caption{Single GPU profiling (RTX 4090, Llama-3-8B + LoRA, bfloat16).}
\label{tab:gpu1}
\begin{tabular}{lccc}
\toprule
Scenario & VRAM Peak (GB) & Latency/step (ms) & Omega Cost (ms) \\
\midrule
Baseline (LoRA)  & 21.98 & 251.7 & 0.0 \\
Omega-S $K{=}1$  & 21.99 & 348.4 & 41.4 \\
Omega-S $K{=}10$ & 21.99 & 261.1 & \textbf{3.5} \\
\bottomrule
\end{tabular}
\end{table}

VRAM overhead: $+13$\,MB ($+0.06\%$). At $K{=}10$, latency overhead is
$+3.7\%$ with an isolated Omega cost of 3.5\,ms per application.

\begin{table}[ht]
\centering
\caption{Two-GPU FSDP profiling (2$\times$ RTX 4090 PCIe, Llama-3-8B + LoRA).}
\label{tab:fsdp}
\begin{tabular}{lccc}
\toprule
Scenario & VRAM/GPU (GB) & Latency/step (ms) & Omega Cost (ms) \\
\midrule
Baseline FSDP    & 17.18 & 14,343 & 0.0 \\
Omega-S $K{=}1$  & 17.18 & 14,273 & 81.3 \\
Omega-S $K{=}10$ & 17.18 & 14,565 & \textbf{8.1} \\
\bottomrule
\end{tabular}
\end{table}

The relative overhead of Omega-S at $K{=}10$ is $+1.5\%$, lower than in
the single-GPU setting, because LoRA adapter weights are not sharded by
FSDP, meaning Omega-S computation requires \textbf{zero all-gather
communication}. \emph{The overhead does not scale with the number of GPUs.}

Note: the high absolute latency in Table~\ref{tab:fsdp} reflects PCIe
interconnect limitations (no NVLink) in the experimental pod. Relative
overhead figures are hardware-independent.

% ============================================================
\subsection{What Omega-S Is and Is Not}

Omega-S is a structural regulariser: it acts on the weight connectivity
graph $A=WW^\top$, not on the curvature of the loss landscape, so it is
complementary to SAM (a full characterisation of their interaction is left
to future work). We are, however, precise about \emph{how} it acts. Its
objective is topological by construction (built from $\operatorname{Tr}(A^3)$,
whose node-wise decomposition carries non-spectral information, Section~3),
but the direct measurement of Section~\ref{sec:mechanism} shows that, as
formulated, the penalty acts entirely through the degree-variance term: the
clustering channel is saturated by the logistic map building $A$ and does
not move ($\Delta C\approx0$ across all ten seeds). We therefore describe
Omega-S, in its present form, as a \emph{degree-variance regulariser with a
topological objective whose clustering channel is currently inert}, rather
than as a working topological method. Whether the effect is fully reducible
to equalising weight-row norms is not settled: degree variance correlates
$r\approx0.6$ with row-norm variance here, leaving a substantial remainder,
and a control swept on its own grid does not reproduce the effect: it fails to
beat no regularisation reliably ($6/10$) while Omega-S exceeds it on $8/10$
(one-sided $p=0.055$). That is evidence against full reducibility, short of
conventional significance at this seed count. The saturation is a design artefact of the
$\sigma(|WW^\top|)$ construction, not an intrinsic limit of the approach; the
reformulation in the Appendix would restore the clustering channel and let a
genuinely topology-driven variant be evaluated.

\subsection{Structured Sparsity as an Emergent Property}

Omega-S alone does not produce structured sparsity (Table~\ref{tab:sweep},
$\lambda{=}0$ row). Structured sparsity emerges only in combination with
group-lasso regularisation. In the MLP pruning experiments Omega-S shifts
\emph{which} neurons become structurally redundant while group-lasso
provides the explicit pressure to collapse them to zero. We describe this as
degree-structure-guided group sparsity; given the mechanism measurement
above, we avoid attributing it specifically to ``topological centrality''
and note that the marginal contribution of Omega-S over group-lasso alone in
this pruning setting remains to be isolated.

\subsection{Biological Analogy and Its Limits}

The microbial network analogy is productive but imperfect. Biological
scale-free networks emerge from evolutionary pressure over millions of
generations; neural weight graphs emerge from gradient descent over thousands
of steps. The analogy provides a structural intuition, not a mathematical
equivalence: systems that process and route information efficiently tend to
avoid extreme connectivity concentration. The empirical validation of this
intuition constitutes the contribution of this paper; the biological framing
provides motivation and vocabulary, while the mathematical and empirical
results stand independently.

\subsection{Broader Applications}
\label{sec:applications}

Beyond structural pruning and sequential fine-tuning, Omega-S opens
several promising research directions.

In \textbf{model quantisation}, weight monopolies often correlate with
activation outliers, a primary bottleneck for sub-8-bit post-training
quantisation that motivates outlier-handling mechanisms such as
\texttt{LLM.int8()}~\cite{dettmers2022} and SmoothQuant~\cite{xiao2023}. By
redistributing weight density, Omega-S might yield networks that are
inherently more quantisation-friendly. We flag this as a hypothesis rather
than a finding: in preliminary probing at small scale we observed the
expected reduction in degree variance but no corresponding improvement in
post-quantisation accuracy, with weight decay performing comparably or
better. Whether the effect appears at LLM scale, where outlier structure is
far more pronounced, remains open.

In \textbf{Federated Learning}, applying a unified topological prior across
decentralised nodes could mitigate structural divergence of local models
trained on heterogeneous data, potentially improving weight aggregation
stability. The absence of inter-GPU communication cost demonstrated in
Section~4.5 suggests negligible overhead even in bandwidth-constrained
settings.

In \textbf{Mixture of Experts (MoE)} architectures, the framework could
penalise routing monopolies, preventing expert collapse and ensuring a
more balanced token-to-expert assignment distribution.

The formulation extends beyond artificial neural networks. In
\textbf{computational biology}, applying Omega-S constraints to graph
neural networks predicting biological pathways could yield representations
better aligned with the natural robustness of living systems. In
\textbf{quantitative finance}, the penalty could regularise financial ML
models to inherently penalise over-reliance on clustered economic nodes,
yielding more robust portfolio optimisation and risk forecasting.

These interdisciplinary applications underscore the fundamental nature of
topological regularisation across any domain where network structure
dictates system resilience, a principle observed in global soil
microbiomes~\cite{ortiz2021,saati2026,biome_patent} long before it was
formalised in the context of artificial neural networks.

\section{Declaration of LLM Usage}
% ============================================================

Large language models were used as writing and coding assistants during the
preparation of this manuscript. Specifically, they were used to improve
language clarity and presentation, to assist with coding and debugging for
the numerical experiments, and to help identify and formalise related work.
All technical ideas, theoretical results, experimental design and final
interpretations were developed, verified and approved by the author.

\section{Additional tables}

\begin{table}[ht]
\centering
\caption{Per-seed retention, ten-seed experiment. $\Delta$ columns are
Omega-S (log-ratio) minus each arm, in percentage points. Omega-S wins $8/10$
against no regulariser, $9/10$ against weight decay and $8/10$ against EWC.
The seeds it loses are informative in both cases: against EWC (123, 3033) they
are the two where EWC posts its highest retention, and against the
no-regulariser arm (123, 5055) they are the first and third highest of that
column.}
\label{tab:perseed10}
\small
\begin{tabular}{lcccccccc}
\toprule
Seed & None & W.\ decay & EWC & Omega-S & $\Delta$ none & $\Delta$ WD & $\Delta$ EWC \\
\midrule
42   & 66.7\% & 64.4\% & 68.3\% & 62.3\%  & $-$4.4  & $-$2.2  & $-$6.0 \\
123  & 85.1\% & 69.2\% & 93.6\% & 87.8\%  & $+$2.7  & $+$18.5  & $-$5.9 \\
456  & 69.6\% & 67.4\% & 77.1\% & 91.1\%  & $+$21.5  & $+$23.7  & $+$14.0 \\
789  & 51.2\% & 58.1\% & 55.3\% & 105.6\%  & $+$54.3  & $+$47.4  & $+$50.2 \\
1011 & 32.6\% & 44.4\% & 40.8\% & 70.8\%  & $+$38.2  & $+$26.4  & $+$30.0 \\
2022 & 69.2\% & 69.6\% & 71.4\% & 82.5\%  & $+$13.3  & $+$12.9  & $+$11.1 \\
3033 & 61.9\% & 64.4\% & 90.7\% & 83.7\%  & $+$21.8  & $+$19.2  & $-$7.0 \\
4044 & 68.2\% & 69.6\% & 87.8\% & 85.5\%  & $+$17.3  & $+$15.9  & $-$2.3 \\
5055 & 72.1\% & 63.6\% & 60.0\% & 74.1\%  & $+$2.0  & $+$10.4  & $+$14.1 \\
6066 & 52.2\% & 47.9\% & 46.0\% & 97.6\%  & $+$45.4  & $+$49.7  & $+$51.6 \\
\midrule
\textbf{Mean} & 62.9\% & 61.9\% & 69.1\% & \textbf{84.1\%} &
\textbf{$+$21.2} & \textbf{$+$22.2} & \textbf{$+$15.0} \\
\bottomrule
\end{tabular}
\end{table}

\begin{table}[ht]
\centering
\caption{Group-lasso $\lambda$ sweep (Omega-S + GL combined).}
\label{tab:sweep}
\begin{tabular}{lccccc}
\toprule
GL $\lambda$ & Accuracy & Sparsity & Dead Cols & $\Delta$FLOPs & Structured \\
\midrule
0.0 (Omega-S only) & 95.45\% & 3.17\% & 0.00\% & 0.00\% & $\times$ \\
$3\times10^{-3}$   & 95.52\% & 9.66\% & 5.65\% & 3.40\% & \checkmark \\
$1\times10^{-2}$   & 94.89\% & 48.22\% & 31.59\% & 26.32\% & \checkmark \\
\bottomrule
\end{tabular}
\end{table}

\vspace{1em}
\noindent\textit{Correspondence: Alberto Acedo (Biome Makers Inc.). The author declares no
competing financial interests beyond the pending patent application
referenced herein.}

\appendix

\section{A contrast-preserving reformulation of the clustering term}
\label{app:reform}

Section~\ref{sec:mechanism} showed that, as implemented, the clustering
channel $C$ is inert: the pseudo-adjacency $A=\sigma(|WW^\top|)$ compresses
all affinities into $[0.5,1)$, so $A$ is near an all-ones matrix and the
normalised clustering $\operatorname{Tr}(A^3)/\lVert A\rVert_F^3$ sits at its
ceiling ($\approx0.9997$) irrespective of weight structure. The penalty can
therefore only act through the degree-variance term $\mathrm{Coex}$. This
appendix sketches how to restore an informative clustering channel, so that
a genuinely clustering-driven variant of Omega-S can be tested. We present
it as an open direction for the community, not as a result of this paper.

The problem is the saturating map, not $\operatorname{Tr}(A^3)$ itself. Any
of the following keeps $C$ responsive to structure:

\paragraph{(i) Temperature-scaled affinity.} Replace
$A=\sigma(|WW^\top|)$ with $A_\tau=\sigma\!\big(|WW^\top|/\tau - b\big)$,
choosing the temperature $\tau$ and bias $b$ so that the affinities span a
usable range of $[0,1]$ rather than collapsing to $[0.5,1)$. Concretely,
set $b$ to the median of $|WW^\top|/\tau$ so that roughly half the entries
fall below $0.5$; then $A_\tau$ has genuine contrast and
$\operatorname{Tr}(A_\tau^3)$ varies with the actual triadic structure. The
objective and its Hutchinson estimator are otherwise unchanged.

\paragraph{(ii) Thresholded graph.} Form a sparse graph
$A_\theta = \mathbf{1}\{|WW^\top| > \theta\}$ (or a smooth surrogate
$\sigma(\kappa(|WW^\top|-\theta))$ with large $\kappa$), with $\theta$ a
per-layer quantile of $|WW^\top|$. Clustering on a thresholded graph is the
standard network-science definition and is not pinned near $1$; its third
moment then measures true triadic closure. The trade-off is that the hard
threshold must be relaxed (via the smooth surrogate) to keep the penalty
differentiable.

\paragraph{(iii) Rank-normalised affinity.} Map $|WW^\top|$ through its
empirical CDF so that $A$ is uniform on $[0,1]$ by construction. This
removes scale and saturation effects entirely and makes $C$ depend only on
the \emph{ordering} of affinities, i.e.\ on graph topology rather than
magnitude.

Under any of these, one can define a clustering-only arm and a
degree-variance-only arm (the row-norm / $\mathrm{Coex}$ control of the main
text, re-tuned to remain a functional learner). One point must be settled
before such an arm is run, and we flag it rather than resolve it: \emph{nothing
measured here determines the sign of the clustering term}. The objective as
implemented minimises $\operatorname{Tr}(A^3)$; the resilience index from which
it derives~\cite{acedo2026fsri} carries the clustering coefficient in its
numerator and therefore rewards raising it; and the measurement of
Section~\ref{sec:mechanism} does not arbitrate between the two, because $C$ did
not move at all. A clustering-only arm should accordingly test \emph{both}
directions, penalising $+\operatorname{Tr}(A^3)/\lVert A\rVert_F^3$ and
$-\operatorname{Tr}(A^3)/\lVert A\rVert_F^3$, rather than assuming one. An
earlier draft of this appendix specified only the second, which was an
unexamined inheritance from the index rather than a considered choice. Running both,
plus full Omega-S, on the ten-seed protocol would decompose the retention
gain into a clustering component and a degree-variance component, and would
settle whether the topological channel (inert here only because of
$\sigma$) contributes once it is allowed to. We conjectured, in an earlier version of this
appendix, that a contrast-preserving $C$ would add retention on the seeds
where the weight graph departs most from a degree-only description (the
$\approx40\%$ of degree variance not captured by row norms).

\paragraph{Outcome.} That experiment has since been run and the conjecture
does not hold. Section~\ref{sec:cosine} reports it in full: a row-wise cosine
affinity removes the saturation, makes the clustering channel steerable during
training, and then loses to the original formulation on all ten seeds
($p=0.002$), with a clustering-only arm worse still. We leave this appendix
in place because the diagnosis it contains is correct and because the
reasoning that led to the reformulation is what made the test possible; the
conjecture at its close is what the evidence overturns.


\begin{thebibliography}{99}

\bibitem{ortiz2021}
Ortiz-\'Alvarez R, Ortega-Arranz H, Ontiveros VJ, de Celis M, Ravarani C,
Acedo A, Belda I.
Network Properties of Local Fungal Communities Reveal the Anthropogenic
Disturbance Consequences of Farming Practices in Vineyard Soils.
\textit{mSystems}. 2021;6(3):e00344-21.
doi: \href{https://doi.org/10.1128/mSystems.00344-21}{10.1128/mSystems.00344-21}.

\bibitem{saati2026}
Saati-Santamar\'ia Z, P\'erez-Gorj\'on S, Abel-Schaad D, et al.
Soil Microbial Diversity and Network Organization Respond to Land Use and
Agricultural Inputs Worldwide.
\textit{Global Change Biology}. 2026;32(7):e70984.
doi: \href{https://doi.org/10.1111/gcb.70984}{10.1111/gcb.70984}.

\bibitem{acedo2022indices}
Acedo A, Ortega-Arranz H, Almonacid D, Ferrero A (2022).
Methods and systems for generating and applying agronomic indices from
microbiome-derived parameters.
US Patent Application Publication No.\ US~2022/0268756~A1
(Appl.\ No.\ 17/665,332, filed 4 February 2022), Biome Makers Inc.

\bibitem{biome_patent}
Ortiz-Alvarez R, Ferrero A, Imam N, Acedo A, Belda-Aguilar I, Almonacid D
(2021).
Methods and systems for evaluating ecological disturbance of an agricultural
microbiome based upon network properties of organism communities.
WIPO Publication No.\ WO~2021/119528~A1 (PCT/US2020/064668), Biome Makers Inc.

\bibitem{foret2021}
Foret P, Kleiner A, Mobahi H, Neyshabur B.
Sharpness-Aware Minimization for Efficiently Improving Generalization.
\textit{ICLR}. 2021.

\bibitem{tam2020}
Tam E, Dunson D.
Fiedler Regularization: Learning Neural Networks with Graph Sparsity.
\textit{ICML}. 2020. arXiv:2003.00992.
Extended version: Spectral Gap Regularization of Neural Networks.
arXiv:2304.03096. 2023.

\bibitem{hutchinson1990}
Hutchinson MF.
A stochastic estimator of the trace of the influence matrix for Laplacian
smoothing splines.
\textit{Communications in Statistics --- Simulation and Computation}.
1990;19(2):433--450.

\bibitem{loshchilov2019}
Loshchilov I, Hutter F.
Decoupled Weight Decay Regularization.
\textit{ICLR}. 2019.

\bibitem{hu2022}
Hu EJ, Shen Y, Wallis P, Allen-Zhu Z, Li Y, Wang S, et al.
LoRA: Low-Rank Adaptation of Large Language Models.
\textit{ICLR}. 2022.

\bibitem{zhao2023}
Zhao Y, Gu A, Varma R, Luo L, Huang CC, Xu M, et al.
PyTorch FSDP: Experiences on Scaling Fully Sharded Data Parallel.
\textit{VLDB}. 2023.

\bibitem{yuan2006}
Yuan M, Lin Y.
Model selection and estimation in regression with grouped variables.
\textit{Journal of the Royal Statistical Society: Series B}.
2006;68(1):49--67.

\bibitem{barabasi1999}
Barab\'asi AL, Albert R.
Emergence of scaling in random networks.
\textit{Science}. 1999;286(5439):509--512.

\bibitem{meyer2021}
Meyer RA, Musco C, Musco C, Woodruff DP.
Hutch++: Optimal Stochastic Trace Estimation.
\textit{SIAM Symposium on Simplicity in Algorithms}. 2021.

\bibitem{yin2024owl}
Yin L, Wu J, Zhang Z, et al. (2024).
Outlier Weighted Layerwise Sparsity (OWL): A Missing Secret Sauce for
Pruning LLMs to High Sparsity.
\textit{arXiv:2310.05175}.

\bibitem{meta2024}
Llama Team, AI @ Meta (2024).
The Llama 3 herd of models.
\emph{arXiv:2407.21783}.

\bibitem{chen2021evaluating}
Chen M, Tworek J, Jun H, Yuan Q, Pinto HPO, Kaplan J, Edwards H, Burda Y,
Joseph N, Brockman G, et al.\ (2021).
Evaluating large language models trained on code.
\emph{arXiv:2107.03374}.

\bibitem{codealpaca2023}
Chaudhary S (2023).
Code Alpaca: an instruction-following LLaMA model for code generation.
\url{https://github.com/sahil280114/codealpaca}.

\bibitem{merity2017pointer}
Merity S, Xiong C, Bradbury J, Socher R (2017).
Pointer sentinel mixture models.
\emph{International Conference on Learning Representations}.

\bibitem{dettmers2022}
Dettmers T, Lewis M, Belkada Y, Zettlemoyer L (2022).
LLM.int8(): 8-bit matrix multiplication for transformers at scale.
\emph{Advances in Neural Information Processing Systems} 35:30318--30332.

\bibitem{xiao2023}
Xiao G, Lin J, Seznec M, Wu H, Demouth J, Han S (2023).
SmoothQuant: accurate and efficient post-training quantization for large
language models.
\emph{International Conference on Machine Learning}, PMLR 202:38087--38099.

\bibitem{acedo2026fsri}
Acedo A. (2026).
The Functional Symbiotic Resilience Index: topological entropy, Wasserstein
curvature bounds and non-equilibrium thermodynamics of complex networks. A
framework linking
network structure, non-equilibrium thermodynamics and discrete curvature.
\emph{Preprint}.

\bibitem{lu2026muonogd}
Lu B, Deng Z, Zhang R, Hu B, Zhao Y, Tian Y, Mou C, Lin G, Li X (2026).
Muon-OGD: Muon-based Spectral Orthogonal Gradient Projection for LLM
Continual Learning.
\emph{arXiv:2605.08949}.

\bibitem{srmoe2026}
Delibasoglu I (2026).
Spectral Manifold Regularization for Stable and Modular Routing in Deep
MoE Architectures.
\emph{arXiv:2601.03889}.

\bibitem{cflowrank2026}
Hidekel IN, Raviv D (2026).
Catastrophic Forgetting is Low-Rank: A Function-Space Theory for Continual
Adaptation.
\emph{arXiv:2606.18024}.

\bibitem{morin2020}
Morin M, Willetts M (2020).
Non-determinism in TensorFlow ResNets.
\emph{arXiv:2001.11396}.

\bibitem{nagarajan2018}
Nagarajan P, Warnell G, Stone P (2018).
Deterministic Implementations for Reproducibility in Deep Reinforcement
Learning.
\emph{arXiv:1809.05676}.

\bibitem{pinto2021}
Pinto A F, et al.\ (2021).
On the reproducibility of neural network predictions.
\emph{arXiv:2105.05482}.

\bibitem{steele2026}
Steele B (2026).
Subspace Geometry Governs Catastrophic Forgetting in Low-Rank Adaptation.
\emph{arXiv:2603.02224}.

\bibitem{ning2026sae}
Ning E, Xue W, Lou D, Guo Y (2026).
From Weights to Features: SAE-Guided Activation Regularization for LLM
Continual Learning.
\emph{arXiv:2606.26629}.

\bibitem{lundstrom2026}
Laitinen-Fredriksson Lundstrom-Imanov GOY (2026).
Mechanistic Analysis of Catastrophic Forgetting in Large Language Models
During Continual Fine-tuning.
\emph{arXiv:2601.18699}.

\bibitem{kirkpatrick2017}
Kirkpatrick J, Pascanu R, Rabinowitz N, Veness J, Desjardins G, Rusu AA,
Milan K, Quan J, Ramalho T, Grabska-Barwinska A, et al.\ (2017).
Overcoming catastrophic forgetting in neural networks.
\emph{PNAS} 114(13):3521--3526.

\bibitem{li2017lwf}
Li Z, Hoiem D (2017).
Learning without forgetting.
\emph{IEEE TPAMI} 40(12):2935--2947.

\bibitem{farajtabar2020}
Farajtabar M, Azizan N, Mott A, Li A (2020).
Orthogonal gradient descent for continual learning.
\emph{AISTATS}, PMLR 108:3762--3773.

\bibitem{wang2023olora}
Wang X, Chen T, Ge Q, Xia H, Bao R, Zheng R, Zhang Q, Gui T, Huang X (2023).
Orthogonal subspace learning for language model continual learning.
\emph{Findings of EMNLP 2023}, 10658--10671.

\bibitem{nayak2025}
Nayak NS, Killamsetty K, Han L, Bhandwaldar A, Chanda P, Xu K, Wang H,
Pareja A, Silkin O, Eyceoz M, et al.\ (2025).
Sculpting subspaces: Constrained full fine-tuning in LLMs for continual
learning.
\emph{arXiv:2504.07097}.

\bibitem{wang2023trace}
Wang X, Zhang Y, Chen T, Gao S, Jin S, Yang X, Xi Z, Zheng R, Zou Y,
Gui T, et al.\ (2023).
TRACE: A comprehensive benchmark for continual learning in large language
models.
\emph{arXiv:2310.06762}.

\bibitem{lu2024alpha}
Lu Y, et al. (2024).
AlphaPruning: Using Heavy-Tailed Self Regularization Theory for Improved
Layer-wise Pruning of Large Language Models.
\textit{arXiv:2410.10912}.

\bibitem{svnup2025}
Anonymous. (2025).
Efficient Shapley Value-based Non-Uniform Pruning of Large Language Models.
\textit{arXiv:2505.01731}.

\bibitem{tang2025}
Tang S, Sieberling O, Kurtic E, Shen Z, Alistarh D. (2025).
DarwinLM: Evolutionary Structured Pruning of Large Language Models.
\textit{arXiv:2502.07780}.

\bibitem{mocanu2016}
Mocanu D C, Mocanu E, Nguyen P H, Gibescu M, Liotta A. (2016).
A Topological Insight into Restricted Boltzmann Machines.
\textit{Machine Learning}, 104, 243--270.

\bibitem{sun2024wanda}
Sun M, Liu Z, Bair A, Kolter J Z. (2024).
A Simple and Effective Pruning Approach for Large Language Models (Wanda).
\textit{ICLR 2024}.

\bibitem{frantar2023sparsegpt}
Frantar E, Alistarh D. (2023).
SparseGPT: Massive Language Models Can Be Accurately Pruned in One Shot.
\textit{ICML 2023}.

\end{thebibliography}
\end{document}